\documentclass[draft]{agujournal2019}
\usepackage{url} %this package should fix any errors with URLs in refs.
\usepackage[inline]{trackchanges} %for better track changes. finalnew option will compile document with changes incorporated.
\usepackage{makecell}
\usepackage{soul}
\usepackage{algorithm}
\usepackage{algpseudocode}
\usepackage{booktabs}  % For better-looking tables
\usepackage{siunitx}   % For number formatting
\usepackage{amsmath}   % For advanced math formatting

\usepackage{graphicx}
\usepackage{tabularx}
\usepackage{multirow}
\usepackage{subcaption} %to have subfigures available
\usepackage{placeins}
\draftfalse

\journalname{Journal of Advances in Modeling Earth Systems (JAMES)}

\begin{document}

%%%%%%%%%%%%%%%%%%%%%%%%%%%%%%%%%%%%%%%%%%%%%%%
%  TITLE
%
% (A title should be specific, informative, and brief. Use
% abbreviations only if they are defined in the abstract. Titles that
% start with general keywords then specific terms are optimized in
% searches)
%
%%%%%%%%%%%%%%%%%%%%%%%%%%%%%%%%%%%%%%%%%%%%%%%

% Example: \title{This is a test title}

\title{LUCIE: A lightweight uncoupled climate emulator with long-term stability and physical consistency}

%%%%%%%%%%%%%%%%%%%%%%%%%%%%%%%%%%%%%%%%%%%%%%%
%
%  AUTHORS AND AFFILIATIONS
%
%%%%%%%%%%%%%%%%%%%%%%%%%%%%%%%%%%%%%%%%%%%%%%%

% Authors are individuals who have significantly contributed to the
% research and preparation of the article. Group authors are allowed, if
% each author in the group is separately identified in an appendix.)

% List authors by first name or initial followed by last name and
% separated by commas. Use \affil{} to number affiliations, and
% \thanks{} for author notes.
% Additional author notes should be indicated with \thanks{} (for
% example, for current addresses).

% Example: \authors{A. B. Author\affil{1}\thanks{Current address, Antartica}, B. C. Author\affil{2,3}, and D. E.
% Author\affil{3,4}\thanks{Also funded by Monsanto.}}

\authors{Haiwen Guan\affil{1}, Troy Arcomano\affil{2}, Ashesh Chattopadhyay\affil{3}, Romit Maulik\affil{1,2}}

\affiliation{1}{The Pennsylvania State University, State College, PA, USA}
\affiliation{2}{Argonne National Laboratory, Lemont, IL, USA}
\affiliation{3}{University of California, Santa Cruz, Santa Cruz, CA, USA}
% \affiliation{4}{Fourth Affiliation}

%(repeat as many times as is necessary)

% Corresponding author mailing address and e-mail address:

% (include name and email addresses of the corresponding author.  More
% than one corresponding author is allowed in this LaTeX file and for
% publication; but only one corresponding author is allowed in our
% editorial system.)

% Example: \correspondingauthor{First and Last Name}{email@address.edu}

\correspondingauthor{Haiwen Guan}{hzg18@psu.edu}

%%%%%%%%%%%%%%%%%%%%%%%%%%%%%%%%%%%%%%%%%%%%%%%
% KEY POINTS
%%%%%%%%%%%%%%%%%%%%%%%%%%%%%%%%%%%%%%%%%%%%%%%
%  List up to three key points (at least one is required)
%  Key Points summarize the main points and conclusions of the article
%  Each must be 140 characters or fewer with no special characters or punctuation and must be complete sentences

% Example:
% \begin{keypoints}
% \item	List up to three key points (at least one is required)
% \item	Key Points summarize the main points and conclusions of the article
% \item	Each must be 140 characters or fewer with no special characters or punctuation and must be complete sentences
% \end{keypoints}

\begin{keypoints}
\item A long-term stable, physically-consistent, and computationally inexpensive AI-based climate emulator, LUCIE, is proposed.
\item $100$ years of autoregressive simulation with $100$ ensemble members have been executed with LUCIE. 
\item LUCIE captures the long-term climatology, annual cycle, variability, and extremes better than a simplified general circulation model. 
\end{keypoints}

%%%%%%%%%%%%%%%%%%%%%%%%%%%%%%%%%%%%%%%%%%%%%%%
%
%  ABSTRACT and PLAIN LANGUAGE SUMMARY
%
% A good Abstract will begin with a short description of the problem
% being addressed, briefly describe the new data or analyses, then
% briefly states the main conclusion(s) and how they are supported and
% uncertainties.

% The Plain Language Summary should be written for a broad audience,
% including journalists and the science-interested public, that will not have 
% a background in your field.
%
% A Plain Language Summary is required in GRL, JGR: Planets, JGR: Biogeosciences,
% JGR: Oceans, G-Cubed, Reviews of Geophysics, and JAMES.
% see http://sharingscience.agu.org/creating-plain-language-summary/)
%
%%%%%%%%%%%%%%%%%%%%%%%%%%%%%%%%%%%%%%%%%%%%%%%

%% \begin{abstract} starts the second page
%Incorporating TISR as a forcing variable has significantly improved long-term stability and climatology in our model. Additionally, employing orography as a land-sea mask not only enhances the accuracy of surface pressure predictions but also prevents the imprinting of land shapes on all variables that is otherwise caused by incorporating surface pressure into the model. Finally, we conduct a scaling experiment, where we train LUCIE on as less as $2$ years of data, and successively increase the amount of data to $11$ years, still much less than the other state-of-the-art data-driven models. Despite being trained on less data, we demonstrate long-term stability and consistent climatology for $100$ years with biases less than state-of-the-art models. 

\begin{abstract}
We present a lightweight, easy-to-train, low-resolution, fully data-driven climate emulator, LUCIE, that can be trained on as low as $2$ years of $6$-hourly ERA5 data. Unlike most state-of-the-art AI weather models, LUCIE remains stable and physically consistent for $100$ years of autoregressive simulation with $100$ ensemble members. Long-term mean climatology from LUCIE's simulation of temperature, wind, precipitation, and humidity matches that of ERA5 data, along with the variability. We further demonstrate how well extreme weather events and their return periods can be estimated from a large ensemble of long-term simulations. We further discuss an improved training strategy with a hard-constrained first-order integrator to suppress autoregressive error growth, a novel spectral regularization strategy to better capture fine-scale dynamics, and finally an optimization algorithm that enables data-limited (as low as $2$ years of $6$-hourly data) training of the emulator without losing stability and physical consistency. Finally, we provide a scaling experiment to compare the long-term bias of LUCIE with respect to the number of training samples. Importantly, LUCIE is an easy to use model that can be trained in just $2.4$h on a single A-100 GPU, allowing for multiple experiments that can explore important scientific questions that could be answered with large ensembles of long-term simulations, e.g., the impact of different variables on the simulation, dynamic response to external forcing, and estimation of extreme weather events, amongst others.  
\end{abstract}

\section*{Plain Language Summary}

LUCIE is an easy-to-train, data-driven climate model designed to run efficiently with minimal training data. It can be trained on as little as two years of data and remains stable for up to 100 years of climate simulations. LUCIE closely matches real-world, observation-based reanalysis for key weather variables like temperature, wind, and precipitation. The model also performs well at predicting extreme weather events. Proposed improvements in training techniques embedded in LUCIE produce improved climatology. LUCIE is quick to train (2.4 hours on a single GPU) and useful for exploring various scientific questions related to climate through quick training and the ability to run long simulations efficiently.
%%%%%%%%%%%%%%%%%%%%%%%%%%%%%%%%%%%%%%%%%%%%%%%
%
%  BODY TEXT
%
%%%%%%%%%%%%%%%%%%%%%%%%%%%%%%%%%%%%%%%%%%%%%%%

%%% Suggested section heads:
% \section{Introduction}
%
% The main text should start with an introduction. Except for short
% manuscripts (such as comments and replies), the text should be divided
% into sections, each with its own heading.

% Headings should be sentence fragments and do not begin with a
% lowercase letter or number. Examples of good headings are:

% \section{Materials and Methods}
% Here is text on Materials and Methods.
%
% \subsection{A descriptive heading about methods}
% More about Methods.
%
% \section{Data} (Or section title might be a descriptive heading about data)
%
% \section{Results} (Or section title might be a descriptive heading about the
% results)
%
% \section{Conclusions}

\section{Introduction}
Recent years have seen an influx of data-driven models for weather that have been demonstrated to perform on par or even outperform numerical weather prediction models (NWPs)~\cite<>[]{pathak2022fourcastnet,lam2022graphcast,bi2023accurate,price2023gencast,bonev2023spherical,nguyen2024scaling}, at a fraction of the computational cost to run. These models are typically trained on the European Center for Medium Range Weather Forecasts's (ECMWF) reanalysis 5 (ERA5) dataset \cite{Hersbach2020}, and can provide weather forecasts at orders of magnitude lower computational cost than traditional NWPs. Moreover, they provide weather forecasts, on almost all variables (that they are trained on) with accuracies that either match or outperform NWPs. These are purely machine learning-based models and do not impose any physical constraints, yet outperform physics-based NWP for some forecast metrics including root-mean-squared error (RMSE) and anomaly correlation coefficient (ACC). While these models may play a large role in operational weather prediction and climate science in the future, we remark that the physical consistency of these models, often on metrics beyond statistical ones, need to be carefully analyzed before they can become operational.

Recently it has been shown that despite their apparent success, these data-driven models are plagued with several problems. While all of these models are accurate for short time scales, they may become numerically unstable beyond two weeks or show unphysical long-term climate when integrated for several years~\cite{chattopadhyay2023long, bonavita2023limitations}. Several such concerns about the physical consistency of some of these models have been raised in recent studies, e.g., on how these models respond to external forcings (such as equatorial heating) or failure to have the correct physical balances ~\cite{hakim2023dynamical,Massimo2024}. Moreover, these models can be excessively diffusive or anti-diffusive~\cite{keisler2022forecasting, chattopadhyay2023long,bonavita2023limitations}, and the Fourier spectrum of the variables predicted by these models do not correspond to the true spectrum, often being extremely diffusive (if stable for beyond a few weeks), thus casting doubt on how physical these models really are and the true resolution at which these models are predicting~\cite{bonavita2023limitations} despite being trained on $0.25^{\circ}$ data. Another challenge with using these large AI-based models is their computational costs which hampers thorough scientific investigations that may require either re-training the model or fine-tuning it, e.g., GraphCast~\cite{lam2022graphcast} and Pangu~\cite{bi2023accurate} are extremely computationally expensive to \textit{train}, requiring dozens of GPUs or even TPUs (in case of GraphCast), and are not convenient for sandbox experiments that require re-training on other datasets~\cite{subich2024efficient}. The usefulness, sensitivity, and physical consistency of these models can only be verified via extensive experiments involving training with different variables, different training sample sizes, etc., and visualizing how they behave. These are essential to understand different sources of instabilities and other inconsistencies in their prediction, e.g., their poor spectral properties. Moreover, many of these models are limited in the quality of their precipitation forecasts - e.g., Pangu does not predict precipitation and GraphCast produces unrealistic (negative) values. FourCastNet~\cite{pathak2022fourcastnet} does not directly predict precipitation in time and instead uses the rest of the variables to estimate the value of precipitation at each time step, similar to NWP models. Finally, the long-term behavior of these models is crucial for ascertaining whether the absence of physics, especially in terms of resolving the small-scale dynamics, affects the models' behavior in terms of emulating the climate of the Earth's system. A comprehensive analysis of these data-driven models require them to be of low complexity so that it is easy to train, change configurations, add and remove physical constraints, and perform sensitivity analysis to different variables on which they are trained, while also having enough number of variables that make it useful as a weather/climate model.  An example of such a minimal complexity model can be found in \citeA{karlbauer2024advancing} where a small number of atmospheric variables with a simpler U-NET architecture at low resolution was emulated with optimistic performance (albeit with a more complicated data regridding strategy). Despite improved stability as compared to other state-of-the-art models, the model proposed in \citeA{karlbauer2024advancing} only emulated up to $1$ year of the climate.  
 
Long-term stability and physical consistency is essential for a data-driven climate emulator. Assuming a low computational cost associated with emulating a pretrained model, one can inexpensively generate large ensembles of climate simulations, at time scales longer than the time span of the reanalysis data used for training. These ensembles of climate simulations can be used to estimate extreme events and their return periods, especially for gray swans, which may typically have return periods longer than the training period~\cite{emanuel2017assessing}. Furthermore, these ensemble members can be used to estimate the response of the atmosphere to external forcings, which is an essential capability of numerical climate models~\cite{bloch2024green}. Recently, the AI2 Climate Emulator (ACE)~\cite{watt2023ace,Duncan2024} trained a deep learning model on numerical climate model outputs and showed long-term stable simulations by incorporating an \textbf{externally imposed sea surface temperature (SST)} from an expensive coupled ocean-atmosphere numerical simulation. While faithfully reproducing the climate of the reference model and a significant leap towards data-driven climate emulation, we remark that ACE requires access to either climatological or observed SSTs as boundary conditions. 

% \textcolor{red}{Troy: Add a small mention of hybrid modeling and how that is beyond the scope of this work.}

While ACE is a fully data-driven climate emulator, another area of active research is \emph{hybrid modeling}: the incorporation of machine learning into existing conventional physics-based atmospheric models. Hybrid models have successfully demonstrated stable, decades-long simulations with a significant reduction of biases and improved variability~\cite<>[]{watt2021correcting,Arcomano_2022,patel2024exploring,kochkov2024neural}. Hybrid models are currently more accurate than fully data-driven models in terms of reproducing long-term climate, however, are typically more computationally expensive and may suffer from instabilities due to nonlinear interactions between the numerical-based and ML-based components of the model. As of present, this issue has been tough to tackle even when the hybrid model is trained on multi-step forecasts \cite<i.e.>[]{kochkov2024neural}. 

% Another consideration for hybrid modeling is the possibility of needing to inference with legacy Fortran-based code or the need to re-write a dynamical-core into a language that is differentiable and can utilize accelerator-based hardware. The former may require creative solutions to introduce the ML model into the Fortran code leading to computational inefficiencies or bugs (see section 2.2.1 of \citeA{Sanford2023}) and the later may require significant computational engineering and resources. On the contrary, fully data-driven emulators typically require less software, can out of the box utilize accelerated hardware using traditional ML software (e.g. PyTorch and JAX), and can take longer time steps compared to a hybrid model. 

In this paper, we introduce, LUCIE -- a fully data-driven \textbf{L}ightweight \textbf{U}ncoupled (no explicit ocean is there in the model) \textbf{C}l\textbf{I}mate \textbf{E}mulator, which instead of being trained on climate model outputs like ACE, is trained on ERA5 data at T30 resolution. LUCIE, similar to ACE, uses a spherical Fourier neural operator (SFNO) as the backbone of the architecture and addresses some of the major challenges in data-driven emulation, e.g., instability due to \textit{spectral bias} (see \citeA{chattopadhyay2023long}), climatological drifts as seen in GraphCast, Pangu, and FourCastNetv2, especially near the poles, issues with diffusive Fourier spectrum of the predicted variables, and inaccurate estimation of precipitation. Specifically, we demonstrate that it is possible to train LUCIE on just $10$ years of ERA5 data (and as little as $2$ years of ERA5 data) while producing $100$ years of physically-consistent and stable simulations using two training strategies: (a) regular stochastic gradient descent (SGD-) based training which requires a successively larger embedding dimension in the SFNO for stability in the low-data regime and (b) an aggregated gradient method (AGGM-) based training strategy to adapt to low-data regimes without increasing the embedding dimension. Furthermore, LUCIE is  computationally cheap to train, requiring only a single A100 GPU for $2.4$h. Thus, several ablation studies to understand the behavior of the models in terms of representing the spectrum, performance with limited data, as well as large-scale low-frequency variability such as the Madden Julian Oscillation (MJO) have been performed. Finally, since LUCIE can be executed with hundreds of ensembles for a $100$ years, the return period of extreme events in several variables is also estimated. We remark that LUCIE is designed keeping in mind the complexity of the architecture and associated variables that makes it computationally efficient enough to conduct sandbox experiments while simultaneously having enough number of variables to verify the important aspects of climate variability. 

The rest of this paper is organized as follows: (a) in Section \ref{sec:data} we describe the coarse-graining and regridding of the ERA5 data in both the spatial and vertical dimensions; (b) in Section \ref{sec:SFNO}, we introduce the underlying SFNO architecture as the backbone of the emulator, a spectral regularization strategy in the tendency, a first-order integration scheme, and the two training strategies; (c) in Section \ref{sec:results} we analyze the long-term climate statistics (mean, variability, and estimation of extremes) as emulated by LUCIE over $100$ years with $100$ ensemble members compared to ERA5 in terms of accuracy and computational cost; and (d) Section \ref{sec:diss} where we highlight the challenges in LUCIE, and discuss potential future directions in extending it to a 3D atmosphere, and coupling with an ocean emulator. 

% that is trained on coarse-resolution ERA5 data regridded on a T30 Gaussian grid on a \textbf{single A100 GPU for only 2.4h}. Furthermore, while we use total incident solar radiation (TISR) as an input to the model, we do not impose externally obtained SSTs (which require further numerical simulations of expensive climate models). To demonstrate the computational efficiency of LUCIE, we perform $5$ years of autoregressive inference with a $1000$- member ensemble and show that the derived climatology has small biases from the reference ERA5 climatology, faithfully reproduces the annual cycle in surface temperature, and shows an accurate Fourier spectrum of zonal wind over $5$ years. LUCIE is built on a spherical Fourier neural operator (SFNO-)\cite{bonev2023spherical} based architecture similar to ACE, but is extremely computationally efficient to perform rapid prototyping, testing counterfactuals, adding/removing prognostic and diagnostic variables, and performing scientific inquiries into model behaviour. Figure~\ref{fig:cover_pic} shows the framework for LUCIE and the annual cycle of near-surface temperature over $5$ years obtained from autoregressive inference with one ensemble member. It must be noted that while we show climatology over $5$ years with $1000$ ensemble members throughout the paper, \textbf{we have conducted $1000$-years emulation with a single ensemble member of LUCIE with minimal climate drift in order to establish its stability~(Fig.~\ref{fig:1000_year_simulation})} . 

\section {Dataset}
\label{sec:data}

\subsection{ERA5}
In this study, we utilize the ERA5 dataset, regridded to the T30 Gaussian grid and vertically interpolated across specific $\sigma$-levels (see section 2.5 of \citeA{Arcomano_2022} for further details of this dataset). Similar to \citeA{watt2023ace}), we decompose the dataset into three categories of variables, i.e., forcing, prognostic, and diagnostic variables. For prognostic variables, which serve as both inputs and outputs, we use $\sigma_{0.95}$ temperature, $T_{{0.95}}$, specific humidity at $\sigma_{0.34}$, $SH_{{0.34}}$, zonal ($U_{0.34}$) and meridional ($V_{0.34}$) winds, and log of surface pressure, SP. Total incoming solar radiation (TISR) and static orography are used as forcing variables (input only), while the log-transform of precipitation (TP) is treated as a diagnostic variable (output only). Additionally, precipitation is transformed using the expression $\log\left(\text{TP}/\epsilon + 1\right)$. Compared to other data-driven models such as FourCastNet, GraphCast, Pangu, and ACE, LUCIE has far few prognostic variables to be trained on ($\approx 2$GB of data). Table~\ref{tab:variables} provides the list of the variables used to train LUCIE.

\begin{table}[h]
\centering
\begin{tabular}{p{4cm} p{2cm} p{5.5cm}}
\toprule
\text{Prognostic} & \text{Forcing} & \text{Diagnostic} \\
\midrule
Temperature $T_{{0.95}}$  & TISR       & Log of 6-hourly total precipitation \\
Specific humidity $SH_{{0.95}}$   & Orography  &    \\
Zonal wind $U_{0.34}$   &            &    \\
Meridional wind $V_{0.34}$   &            &    \\
Surface pressure SP  &            &    \\
\bottomrule
\end{tabular}
\caption{List of of prognostic, forcing, and diagnostic variables. The subscript for the prognostic variables denotes the corresponding $\sigma$ - level for the variable. For example $T_{{0.95}}$ indicates temperature at $\sigma$ = 0.95 .  }
\label{tab:variables}
\end{table}

\subsection{Simplified Parameterization, primitive-Equation Dynamics (SPEEDY)}

SPEEDY is a spectral transform atmospheric general circulation model, that utilizes simplified, but modern parameterization schemes \cite{molteni2003,Kucharski2006}. We use the standard configuration of Version 41 of the model with a spectral horizontal resolution of T30, with the corresponding Gaussian grid that has a horizontal spatial resolution of 3.75$^\circ \times$3.75$^\circ$. The three-dimensional prognostic variables of the model are the two components of the horizontal wind vector, temperature, and specific humidity defined at eight vertical $\sigma$-levels (0.025, 0.095, 0.20, 0.34, 0.51, 0.685, 0.835, and 0.95), where $\sigma$ is the ratio of pressure to the surface pressure. The single two-dimensional prognostic variable is the natural logarithm of surface pressure and precipitation in the model is diagnosed by parameterizations schemes. The aforementioned ERA5 dataset was regridded from higher resolution to match the state vector of SPEEDY. 

To evaluate LUCIE's ability to simulate the climate, we compare it a 40-year free run of SPEEDY using climatological boundary and forcing conditions. The free run starts from a rest atmospheric state and the first 5 years of the model are discarded. 

\section{Method}
\begin{figure}[!h]
%\vskip 0.2in
\begin{center}
\centerline{\includegraphics[width=\textwidth,scale=1.2]{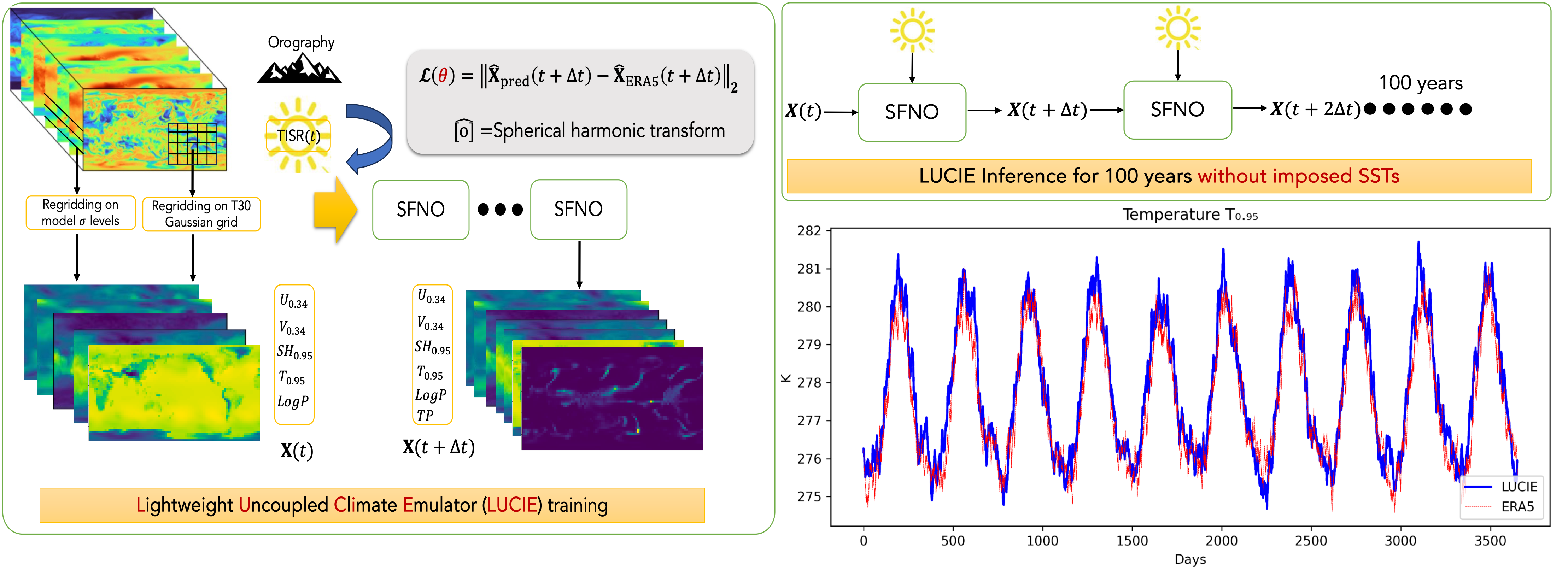}}
\caption{A schematic of LUCIE show the training stage of the model (left panel), inference stage for an individual ensemble member (top right), and global annual mean temperature for $T_{{0.95}}$ for the the first 10 years of inference and ERA5 from 2010-2019 for reference (bottom right).  }
\label{fig:schematic}
\end{center}
% \vskip -0.4in
\end{figure}

In this section, we describe each component of the LUCIE architecture, how it is trained, and how a large number of ensemble members are generated during inference. Fig.~\ref{fig:schematic}  shows the schematic of training and inference parts of LUCIE.

\subsection{SFNO backbone}
\label{sec:SFNO}
In this work, we extend the SFNO architecture as introduced by \cite{bonev2023spherical} to accommodate the spherical representation of Earth's climate data. LUCIE begins with a multilayer perceptron (MLP) featuring a single hidden layer, which projects the input data into a higher-dimensional latent space. To capture spatial information effectively, a trainable, time-independent positional embedding is integrated into the latent variables as biases. These spatially embedded latent variables are then processed through a series of SFNO blocks. Each block conducts Driscol \& Healey \cite{driscoll1994} operations to perform transformations using spherical harmonics. To facilitate more effective learning without loss of information, each SFNO block includes a skip connection that reintroduces input latent variables at the block's output. Finally, the data, transformed through the SFNO blocks, is passed to a single-layer MLP that maps the high-dimensional outputs back to the original physical space. Table~\ref{tab:hyperparameters} refers to hyperparameters used in SFNO architecture training. 

\subsection{Euler integration as a hard constraint}
The global atmosphere, with the states described in section~\ref{sec:data} can be represented by the dynamical system:
\begin{eqnarray}
            \frac{d\mathbf{X}}{dt} = \mathbf{F}\left(\mathbf{X}\right),
\label{eq:dyn_sys}
\end{eqnarray}
where $\mathbf{X}$ are the states of the system and $\mathbf{F}$ is the unknown dynamics of the atmosphere that we emulate with a data-driven model. Most data-driven models auoregressively emulate the discretized version of Eq.~\ref{eq:dyn_sys} with:
\begin{eqnarray}
   \mathbf{X}(t+\Delta t)=\underbrace{\mathbf{X}(t)+ 
      \int_{t}^{t+\Delta t}{ \mathbf{F}\left(\mathbf{X}\left(t\right)\right)}dt}_{\mathcal{M}\left[\mathbf{X}(t),\phi \right]},
      \label{eq:direct_pred}
\end{eqnarray}
where $\mathcal{M}$ is the data-driven model with parameters, $\phi$ that represents the entire right hand size of Eq.~\ref{eq:direct_pred} and $\Delta t$ is the time step used in the data-driven model. In this paper, instead of directly predicting the next time step of the prognostic variables, $\mathbf{X}$, we implement an Euler integration scheme and represent the tendency of the dynamical system in Eq.~\ref{eq:dyn_sys} with the SFNO model:
\begin{eqnarray}
     \mathbf{X^{p}}(t+\Delta t)=\underbrace{\mathbf{X^{p}}(t)+ 
      \int_{t}^{t+\Delta t} \underbrace{\mathbf{F}\left(\mathbf{X^{p}}\left(t\right)\right) dt}_{\mathcal{N[\circ,\theta]}}}_{\mathbf{H[\circ]}}, 
      \label{eq:euler_pde}
\end{eqnarray}

where $\mathcal{N}$ is the SFNO described in section~\ref{sec:SFNO} with parameters, $\theta$, $\mathbf{X^{p}}$ are the prognostic variables, and $\mathbf{H}$ is simply the identity operator. Other integration schemes (for instance higher-order integrators) can also be used and have been shown to be useful in other emulation tasks as well, e.g, see \citeA{chattopadhyay2023long} and \citeA{chattopadhyay2023oceannet} at the cost of higher computational memory during training. The performance of our model, including its stability and long-term physical consistency is significantly improved with this strategy. It must be noted that the integration scheme is applied to only to the prognostic variables, while precipitation is directly diagnosed as an output. In this paper, LUCIE autoregressively integrates the system at a $\Delta t$ of $6$h.

% Rather than mapping input variables—full-field climate variables—directly to the full fields of the next timestep, our model predicts the incremental changes (deltas) of the prognostic variables. This method is proved to be effective in reducing artifacts compared to our experimental models that predict full-field outputs.
% Given that precipitation is a diagnostic variable, predicting the changes in precipitation is not feasible since the model lacks input to learn the initial condition of precipitation. To overcome this, we use a hybrid prediction method. Prognostic variables are predicted as changes (deltas) from their previous states, while precipitation is predicted as a full-field state. We then add the predicted deltas of prognostic variables back to their previous states to reconstruct their full fields. These reconstructed fields are input again into the model for autoregressive prediction of the next steps.

\subsection{Loss function and spectral regularization in the tendency}
In this paper, we use a combined loss function that utilizes the spherical weighted $L_2$ norm in the Legendre-Gauss grid, $\mathbf{L}_{sph} \left(\theta\right)$ with a spectral regularizer, $\mu\left(\theta\right)$,that penalizes the Fourier spectrum of the tendency of the prognostic variables and the full field of precipitation between the prediction and ERA5 at each time step. It should be noted that $\mu\left(\theta\right)$ is computed only over the extra-tropical region of the globe and not over the entire globe. 

The total loss function is:
\begin{eqnarray}
 \mathbf{L}\left(\theta\right) = (1-\alpha)\mathbf{L}_{sph} \left(\theta\right) + \alpha\mu\left(\theta\right),  
\end{eqnarray}

where:
\begin{eqnarray}
\mathbf{L}_{sph} = \Delta \lambda \sum_{t=0}^{t=T}\sum_{i=0}^{n_{\text{lat}}} \sum_{j=0}^{n_{\text{lon}}} \biggr\| \frac{d\mathbf{{X^{p}_{ij}}}}{dt}\left(t\right) -\mathcal{N}\left(\mathbf{X^{p}_{ij}}\left(t\right),\theta\right)\biggr\| _2^2 W_i + \Delta \lambda \sum_{t=0}^{t=T}\sum_{i=0}^{n_{\text{lat}}} \sum_{j=0}^{n_{\text{lon}}} \biggr\| \mathbf{X^{d}_{ij}}\left(t+\Delta t\right)-\mathcal{N}\left(\mathbf{X^{d}_{ij}}\left(t\right),\theta \right)\biggr\|_2^2 W_i,
\end{eqnarray}

and $\mu\left(\theta\right)$ is given by:
\begin{eqnarray}
    \mu\left(\theta\right) = \sum_{t=0}^{t=T}\biggr\| \frac{d\mathbf{\widehat{X^{p}}}}{dt}\left(t\right) -\mathcal{\widehat{N}}\left(\mathbf{X^{p}}\left(t\right),\theta\right)\biggr\|_1 + \sum_{t=0}^{t=T}\biggr\| \widehat{\mathbf{X}^{d}}\left(t+\Delta t\right) - \mathcal{\widehat{N}}\left(\mathbf{X^{d}}\left(t\right),\theta\right)\biggr\|_1.
    \label{eq:spec_reg}
\end{eqnarray}

Here, $\widehat{\left[\right]}$ represents the Fourier spectrum computed as an average over the extra-tropical latitude bands, $T$ is the number of training samples (in time) over which the model is trained, and $\mathbf{X^{d}}$ is the diagnosed variable (precipitation, in our case). $W_i$ is defined by the Legendre-Gauss weights of the latitudes and $\Delta\lambda$ is the longitude interval.

In our study, we find that using a Legendre-Gauss grid with a spherical weighted $L_2$ loss function significantly improves model performance when training on the tendencies of prognostic variables. In contrast, the Mean Squared Error (MSE) loss function shows better effectiveness for full-field predictions. Based on these observations, we apply the spherical weighted $L_2$  loss for the tendency of the predictions of prognostic variables and use the MSE loss for predicting precipitation in full-field states. While incorporating different loss functions within a single model can potentially disrupt the gradient map, our ability to extend the number of training epochs effectively mitigates this issue.

% \subsection{Optimization Strategy}
% Instead of using the standard backpropagation procedure, we implement a novel gradient accumulation method. This technique is different from the typical approach of resetting gradients for each batch. Instead, we accumulate gradients across multiple input batches before updating the optimizer.

% \subsection{Spectral Regularizer}
% To better align the model’s power spectrum with that of ERA5, we have introduced a spectral regularizer [Chattopadhyay et al., 2023] designed to minimize discrepancies between the FFT power spectra of model predictions and actual ERA5 data. Specifically, we apply an FFT transformation to the data's extra-tropical areas and compute the Mean Absolute Error (MAE) loss between these predictions and the truth. This spectral loss, which carries a minor weight, is incorporated during a fine-tuning phase of 50 epochs after the initial training with the previously mentioned loss functions.

\subsection{AGGM training strategy for low-data regime}

Here, we propose the Aggregated Gradient Method (AGGM), a novel training strategy for LUCIE in the low-data regime, e.g., $2$ years of training samples. Unlike traditional optimization methods that reset the gradient of the optimizer after each batch, AGGM resets the optimizer's gradient at the beginning of each training epoch. This approach differs significantly from gradient accumulation methods, which typically averages several gradients before executing a single backpropagation step. Instead, AGGM accumulates the gradients \emph{with backpropagation} for the entire batch. Thus, every backpropagation step in one epoch utilizes all the gradients previous to the current batch of training data. See Table~\ref{sec:aggm_appendix} for details. 

\subsection{Early stopping}

To enhance model convergence, we implement an early stopping mechanism. During training, after 150 epochs, the model performs autoregressive inference every 20 epochs, extending for 2 years. The climatology bias of temperature is calculated using the second year of predictions. This climatology is weighted by a quadrature factor and compared to the ERA5 climatology. If the calculated climatology shows a smaller bias than the previous one, the model saves a checkpoint. Otherwise, the model reverts to the prior checkpoint. If this reversion occurs three times, the model resumes from the final checkpoint and terminates the training process. Once again, we remark that the design decisions utilized here and in previous sections lead to enhanced learning with reduced training data requirements.

\subsection{Ensemble Generation}
To analyze the uncertainty in our model, we generate 100 ensemble members using a weight perturbation technique~\cite{pasc2024pos122}. This technique adds small Gaussian noise to the real component of the weights of all the SFNO spectral filters. In this study, the noise is defined as a Gaussian with zero mean a standard deviation equal to $3e-4$ to avoid unrealistic blow-up during inference. Each ensemble member is then deployed to autoregressively emulate the dynamics of the atmosphere for 100 years using the same out-of-sample initial condition. 

% \ta{need to be precise e.g. are we only adding the real component, do we use any initial condition methods, etc}

\section{Results}
\label{sec:results}

\subsection{Climatology emulated by LUCIE}
\begin{figure}[h]
%\vskip 0.2in
\begin{center}
\centerline{\includegraphics[width=1.1\textwidth]{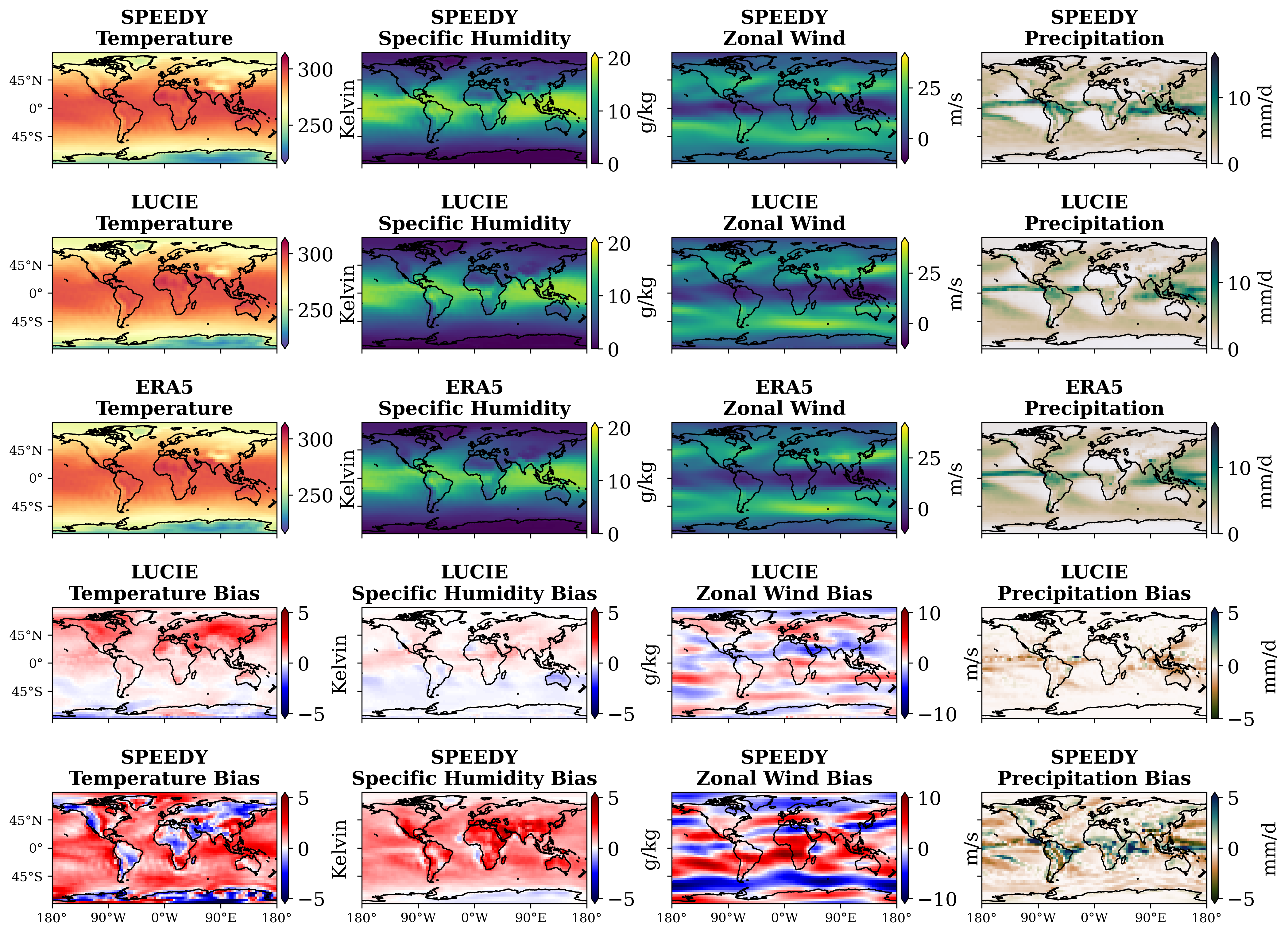}}
\caption{Ensemble mean annual climatology bias of LUCIE for selected variables with respect to the 10 year period of ERA5 and SPEEDY from 2000-2010. LUCIE's climatology is averaged over both the 100 years of simulation and 100 ensemble members. }
\label{fig:clim_bias}
\end{center}
% \vskip -0.4in
\end{figure}

\begin{figure}[!h]
%\vskip 0.2in
\begin{center}
\centerline{\includegraphics[width=0.95\textwidth]{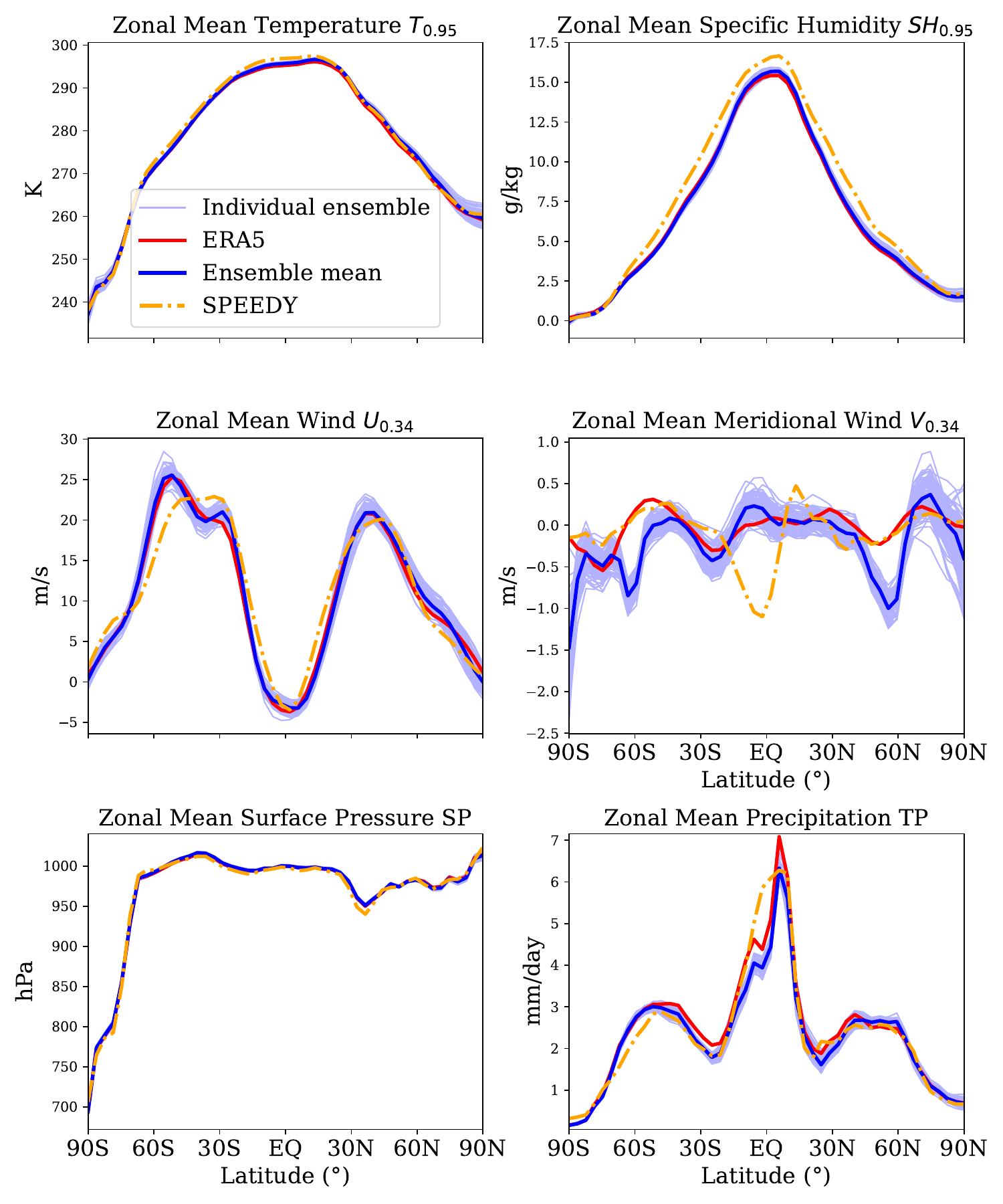}}
\caption{Zonal mean climatology for the 100 LUCIE individual ensemble members (thin blue lines), the LUCIE ensemble mean (thick blue lines), SPEEDY (dashed yellow lines), and ERA5 from 2000 - 2010 (thick red lines). }
\label{fig:clim_ens_zonal}
\end{center}
% \vskip -0.4in
\end{figure}

Our analysis begins with a calculation of the 100-year annual mean climatology across all ensemble members in Fig.~\ref{fig:clim_bias} as well as the zonal mean of individual ensemble members in Fig.~\ref{fig:clim_ens_zonal}. The results of a  singular SPEEDY free run are compared to LUCIE as reference for a coarse, AGCM operating at the same horizontal resolution as LUCIE.  We also report the climatological biases of LUCIE (and SPEEDY), using the ten-year period of ERA5 (2000-2010) as reference, for all variables in Table~\ref{tab:biases}. Overall, in Fig.~\ref{fig:clim_bias}, we can see that LUCIE has low biases in all prognostic variables as well as for precipitation, particularly for a model with such coarse resolution and with only 6 prognostic variables. The mean, minimum, and maximum biases for all of the variables LUCIE predicts or diagnoses are generally on par or better than SPEEDY. In particular, the root-mean-squared biases of LUCIE's climatology are significantly better than SPEEDY, with all variables improving upon SPEEDY by at least a factor of two. We note that SPEEDY's orography is heavily smoothed leading to exaggerated surface pressure biases when compared to ERA5. 

The ensemble-average zonal mean of zonal wind in Fig.~\ref{fig:clim_ens_zonal} shows that there are no artifacts in the poles, a problem that is persistent with all the state-of-the-art AI models. Both SPEEDY and LUCIE are able to reproduce the zonal mean structures for each variable shown, however, LUCIE tends to be closer to ERA5 (e.g., better resolving the two jet structure in the Southern Hemisphere). The zonal mean of precipitation in Fig.~\ref{fig:clim_ens_zonal} emulated by LUCIE shows that the locations of the inter-tropical convergence zones (ITCZ) are correctly captured. We note that the model tends to under-predict annual rainfall, especially over the Maritime Continent. We hypothesize that this is due to the limited number of prognostic variables and the choice to make precipitation a diagnostic variable. In fact, preliminary results suggest that increasing the number of model levels helps eliminate this dry bias. LUCIE has a noticeable warm bias over land for near surface temperature, however, most regions are bounded to  $ \pm 5$ Kelvin with a global average RMSE of $0.73$ Kelvin. The annual climatology for specific humidity matches ERA5 well, with only a slight moist bias collocated with the areas with a warm bias in the Northern Hemisphere and slight dry bias broadly over the Southern Hemisphere, with the largest dry biases over the Australian continent. Specific humidity is a variable that numerical models have particular trouble with due to the effects of parameterizations for the boundary layer and convection/precipitation. This can be seen with SPEEDY, where SPEEDY tends to have significant moisture biases for much of the globe with particularly large biases over land compared to LUCIE. For zonal winds LUCIE is able to capture the general circulation of the atmosphere, particular the polar and subtropical jets. However, our model tends to place the jet stream comparatively northwards in the Northern Hemisphere and has an over active subtropical jet. As with the other variables, LUCIE tends to have smaller biases for most locations in the world compared to SPEEDY. 

We found that incorporating TISR as a forcing variable significantly improves the long-term stability and climatology in our model. Additionally, employing orography as a land-sea mask not only enhances the accuracy of surface pressure predictions, but also prevents the imprinting of land shapes on all variables that is otherwise caused by incorporating surface pressure into the model.

\begin{table}[h!]
\centering
\begin{tabular}{|l|r|r|r|r|}
 \hline
 \cline{2-5}
 & \multicolumn{4}{|c|}{LUCIE (SPEEDY) Climatology} \\
 \hline
 Variable & Min Bias & Max Bias & Mean Bias & RMSE \\
 \hline
 Temperature (K)&   \textbf{-1.544} (-6.812)  & \textbf{2.766} (12.13) & \textbf{0.444} (0.810) & \textbf{0.727} (1.564)\\
 Specific Humidity (g/kg) & -0.979 (\textbf{-0.917}) & \textbf{1.233} (5.641) & \textbf{0.056} (0.932) & \textbf{0.271} (1.389)\\
 Zonal Wind (m/s) & \textbf{-4.626} (-9.760) & \textbf{4.280} (10.69) & \textbf{0.140} (0.385) & \textbf{1.354} (3.582)\\
 Meridional Wind (m/s) & \textbf{-3.548} (-6.156) & \textbf{2.197} (4.343) & -0.177 (\textbf{-0.075}) & \textbf{0.648} (1.458)\\
 Surface Pressure (hPa) & \textbf{-5.949} (-162.5) & \textbf{3.602} (140.7) & \textbf{-0.350} (0.917) & \textbf{1.147} (17.68)\\
 Precipitation (mm/d) & \textbf{-3.020} (-9.411) & \textbf{6.085} (9.795) & -0.152 (\textbf{-0.079}) & \textbf{0.567} (1.490)\\
 \hline
\end{tabular}

\caption{Global minimum, maximum, and mean biases and area-weighted root-mean-squared bias for temperature, specific humidity, and zonal wind of LUCIE and SPEEDY with respect to the 10 year period of ERA5 from 2000 - 2010. }
\label{tab:biases}
\end{table}

\begin{table}[h!]
\centering
\begin{tabular}{c c c c}
\toprule
\textbf{Variables} & \makecell{\textbf{Mean bias} \\ \textbf{without $\mu(\theta)$}} & \makecell{\textbf{Mean bias} \\ \textbf{with $\mu(\theta)$}} & \makecell{\textbf{Percentage} \\ \textbf{improvement}} \\

\midrule
Temperature (K)  & 0.788 & \textbf{0.727} $\pm 1.7\times 10^{-2}$ & 7.7\% \\
Specific Humidity (g/kg)  & 0.328 & \textbf{0.271} $\pm 2.4\times 10^{-9}$ & 17.4\% \\
Zonal Wind (m/s)  & 1.744 & \textbf{1.354} $\pm 4.2\times 10^{-2}$ & 22.3\% \\
Meridional Wind (m/s)  & 0.990 & \textbf{0.648} $\pm 1.7\times 10^{-3}$ & 38.0\% \\
Surface Pressure (hPa)  & 0.153 & \textbf{0.118} $\pm 1.3\times 10^{-2}$ & 22.9\% \\
Precipitation (mm/d)  & 0.611 & \textbf{0.567} $\pm 8.3\times 10^{-5}$ & 7.2\% \\
\bottomrule
\end{tabular}
\caption{$100$-year, $100$-member ensemble averaged climatology bias with and without spectral regularization, $\mu(\theta)$ (see Eq.~\ref{eq:spec_reg}) in LUCIE. }
\label{tab:rmse}
\end{table}\

\subsection{Variability in LUCIE Simulation}
\begin{figure}[!h]
%\vskip 0.2in
\begin{center}
\centerline{\includegraphics[width=0.8\textwidth]{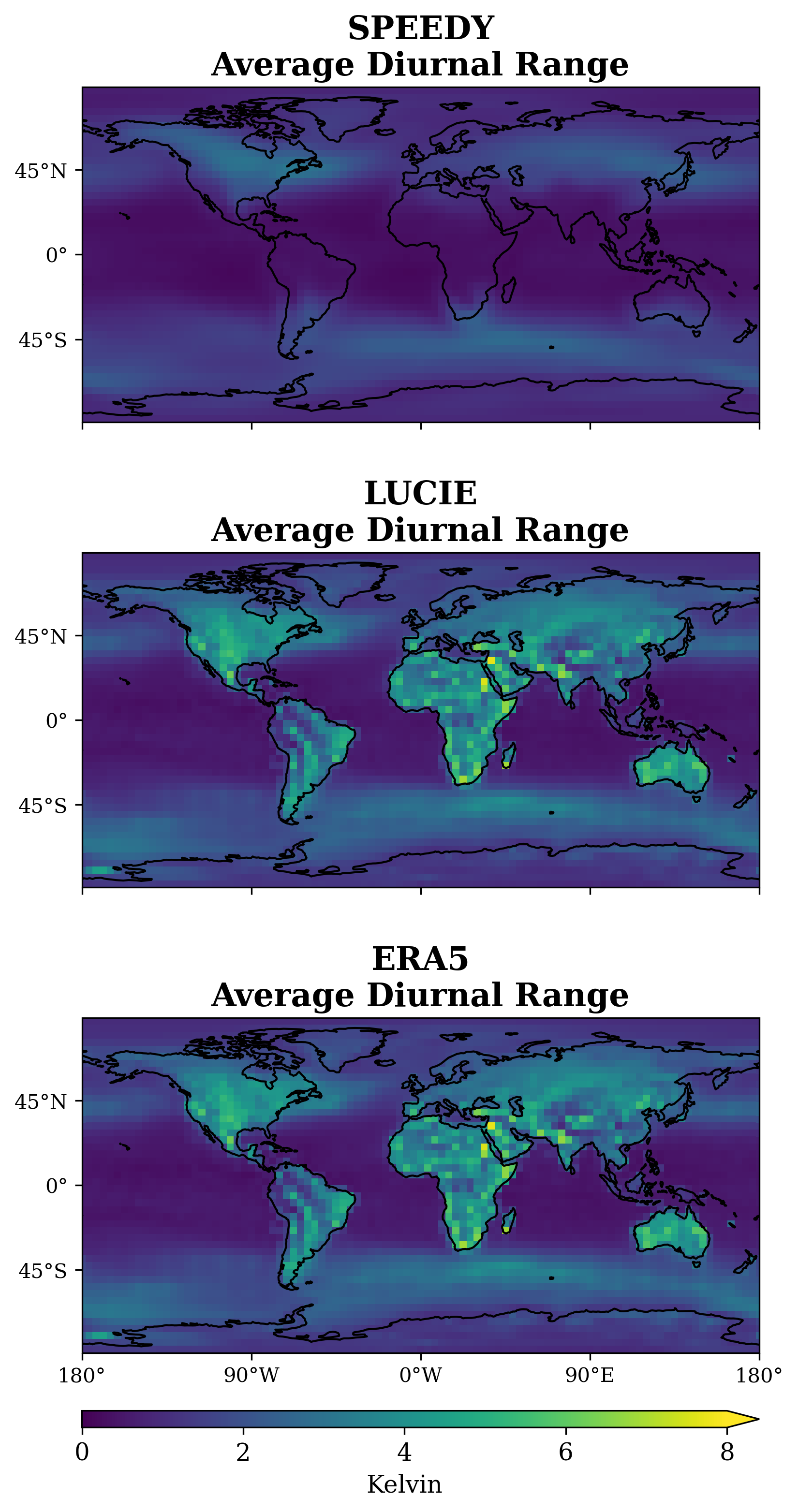}}
\caption{$100$ ensembles of $100$-year averaged diurnal range of $T_{0.95}$ compared to $10$-years average of ERA5 and SPEEDY from 2000 - 2010.}
\label{fig:diurnal_range}
\end{center}
% \vskip -0.4in
\end{figure}

\begin{figure}[!h]
%\vskip 0.2in
\begin{center}
\centerline{\includegraphics[width=0.8\textwidth]{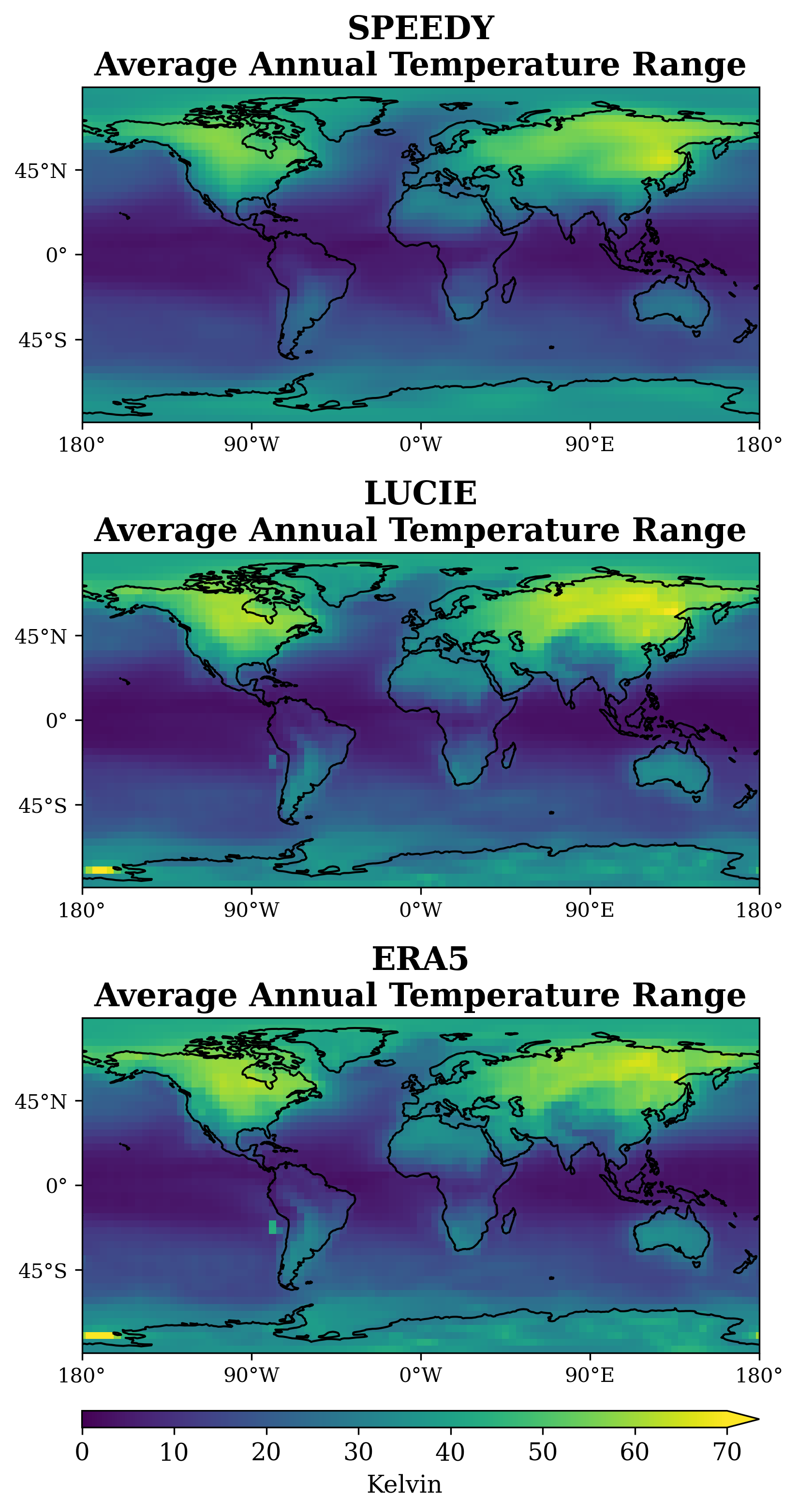}}
\caption{$100$ ensembles of $100$-year averaged annual temperature range compared to $10$-years average of ERA5 and SPEEDY from 2000 - 2010.}
\label{fig:variability}
\end{center}
% \vskip -0.4in
\end{figure}

\begin{figure}[!h]
%\vskip 0.2in
\begin{center}
\centerline{\includegraphics[width=0.8\textwidth]{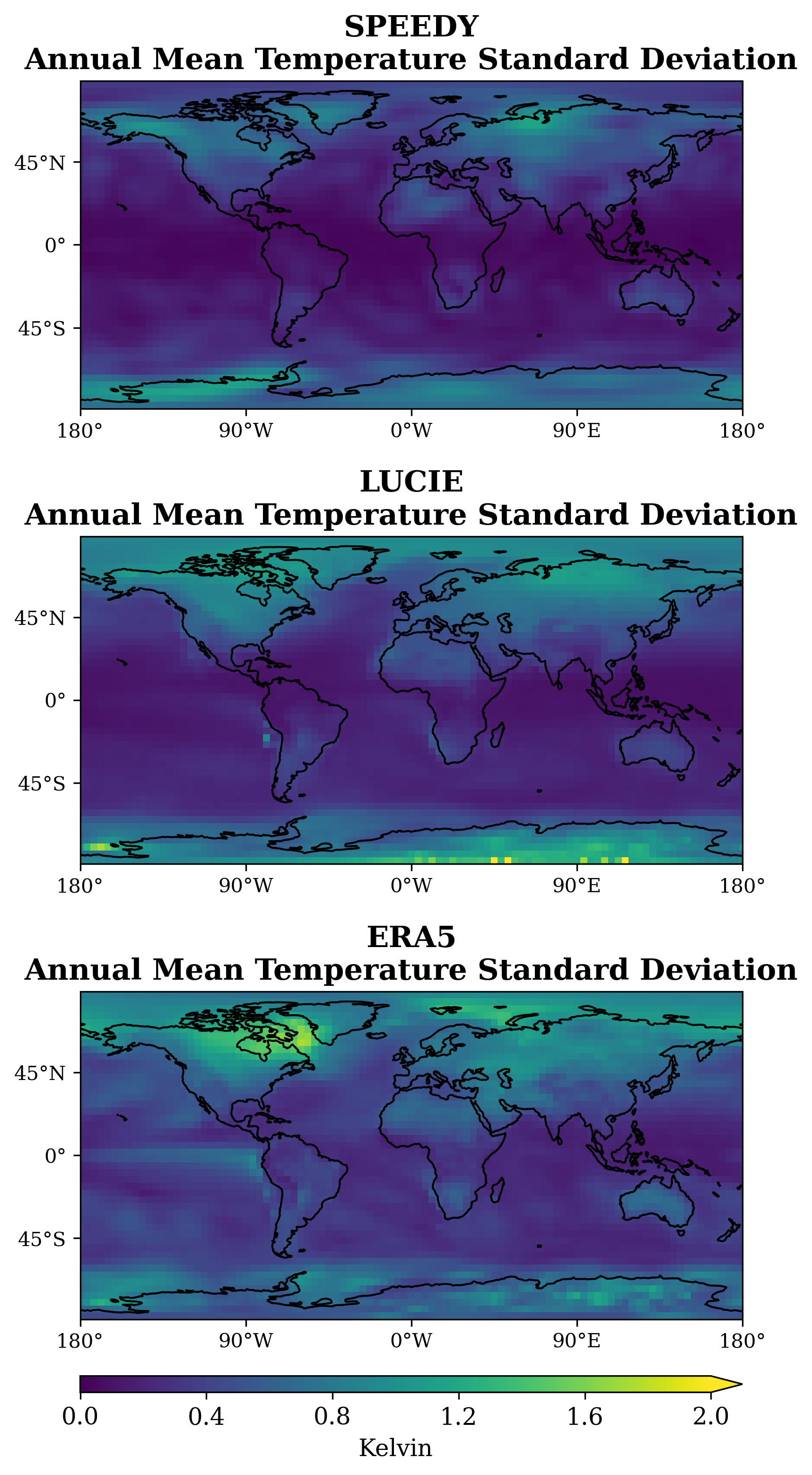}}
\caption{$100$ ensembles of $100$-year averaged standard deviation of annual mean temperature compared to $10$-years average of ERA5 and SPEEDY from 2000 - 2010.}
\label{fig:annual_std}
\end{center}
% \vskip -0.4in
\end{figure}

We investigate the variability in some of the key variables emulated by LUCIE over $100$ years with $100$ ensemble members. Figure~\ref{fig:diurnal_range} shows the $100$-years $100$-ensembles average of the diurnal range of $T_{0.95}$ emulated by LUCIE. The close match between the simulation and ERA5 is promising since it establishes LUCIE's capability to accurately capture the daily variability of temperature over $100$ years. Unsurprisingly, SPEEDY fails to capture the diurnal cycle as one of the simplifications made in the model is to use daily averaged solar fluxes. As also found in \citeA{karlbauer2024advancing}, LUCIE is able to distinguish between dynamics in the planetary boundary layer (PBL) over different vegetation and land-type with no explicit information given about vegetation or other environmental variables, besides orography and total-incoming solar radiation. Both SPEEDY and LUCIE are able to reproduce the general structure of annual temperature range, however, SPEEDY tends underestimate the annual temperature range and is generally more smooth compared to ERA5. LUCIE is also able to distinguish between diurnal cycle over ocean and land, evident by the much smaller diurnal range over the ocean compared to land. Similarly, variability in the annual temperature range is accurately captured by LUCIE as shown in Fig.~\ref{fig:variability}. LUCIE correctly captures the magnitude and locations where the largest annual temperature ranges occur, mainly Siberia and central North America. Lastly, we showcase variability from year to year for the annual mean temperature in Fig.~\ref{fig:annual_std}. Unsurprisingly, without being coupled to ocean and sea-ice models, LUCIE fails to capture the areas with the largest year-to-year variability as seen in ERA5 over the equatorial Pacific (El Niño-Southern Oscillation) and areas influenced by variability in sea-ice extent. This is also reflected in SPEEDY where annually repeating SSTs and sea ice lead to a muted year-to-year variability. 

\begin{figure}[!h]
%\vskip 0.2in
\begin{center}
\centerline{\includegraphics[width=1.1\textwidth]{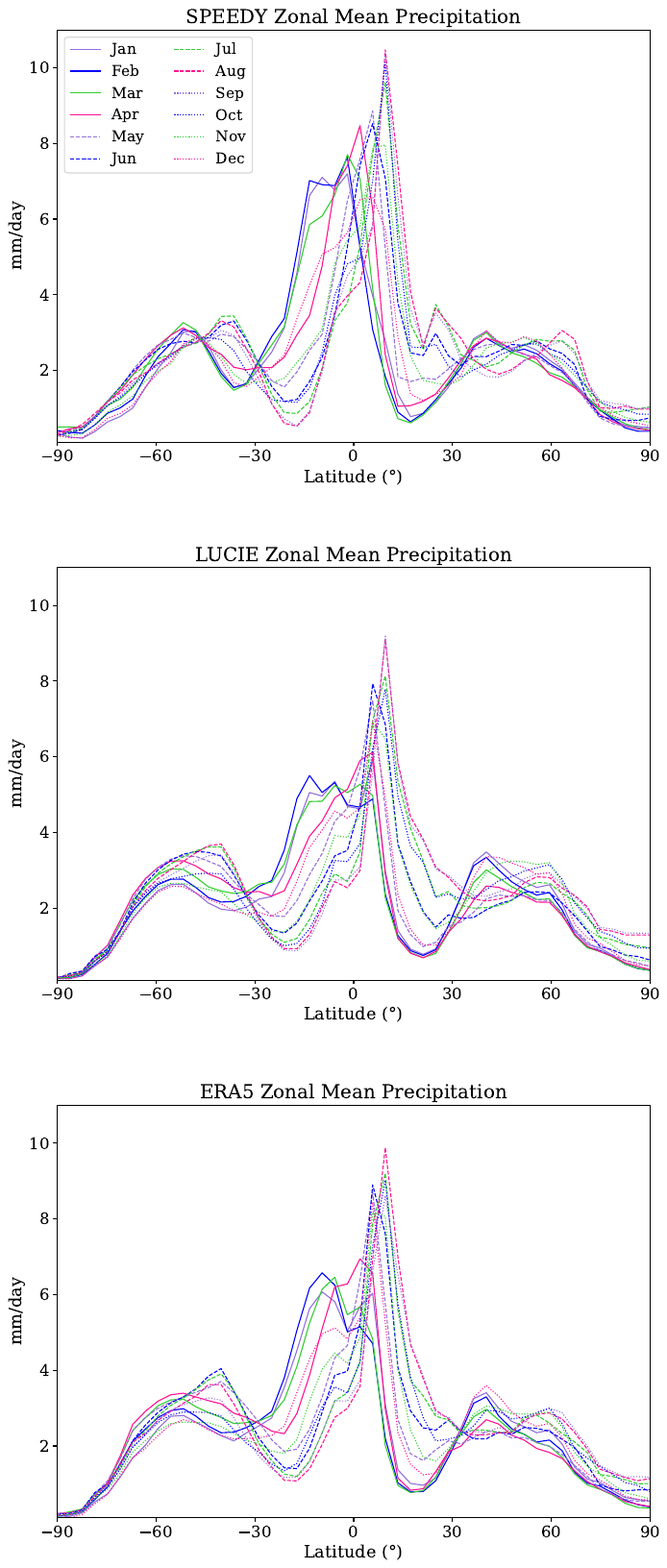}}
\caption{Monthly zonal-mean precipitation of SPEEDY (top panel), LUCIE (mid panel), and ERA5 (bottom panel). Here a $100$-year, $100$-member ensemble average over each month has been computed for LUCIE and a $10$-year average is computed for ERA5 and SPEEDY. }
\label{fig:monthly_precip}
\end{center}
\end{figure}

To assess the variability in precipitation, we investigate the zonal-mean $100$-year and $100$-ensembles average monthly precipitation in Fig.~\ref{fig:monthly_precip}. Generally, LUCIE agrees with ERA5 for the monthly, zonal mean precipitation patterns and how they change over the year. LUCIE is able to capture the movement of the ITCZ through the year (e.g. south of the equator during the boreal winter and the subsequent movement northward during the spring). During the boreal winter, LUCIE tends to have a wider and more south ITCZ compared to ERA5. SPEEDY is also able to capture the month to month changes in the zonal precipitation patterns.

\begin{figure}[!h]
%\vskip 0.2in
\begin{center}
\centerline{\includegraphics[width=1.2\textwidth]{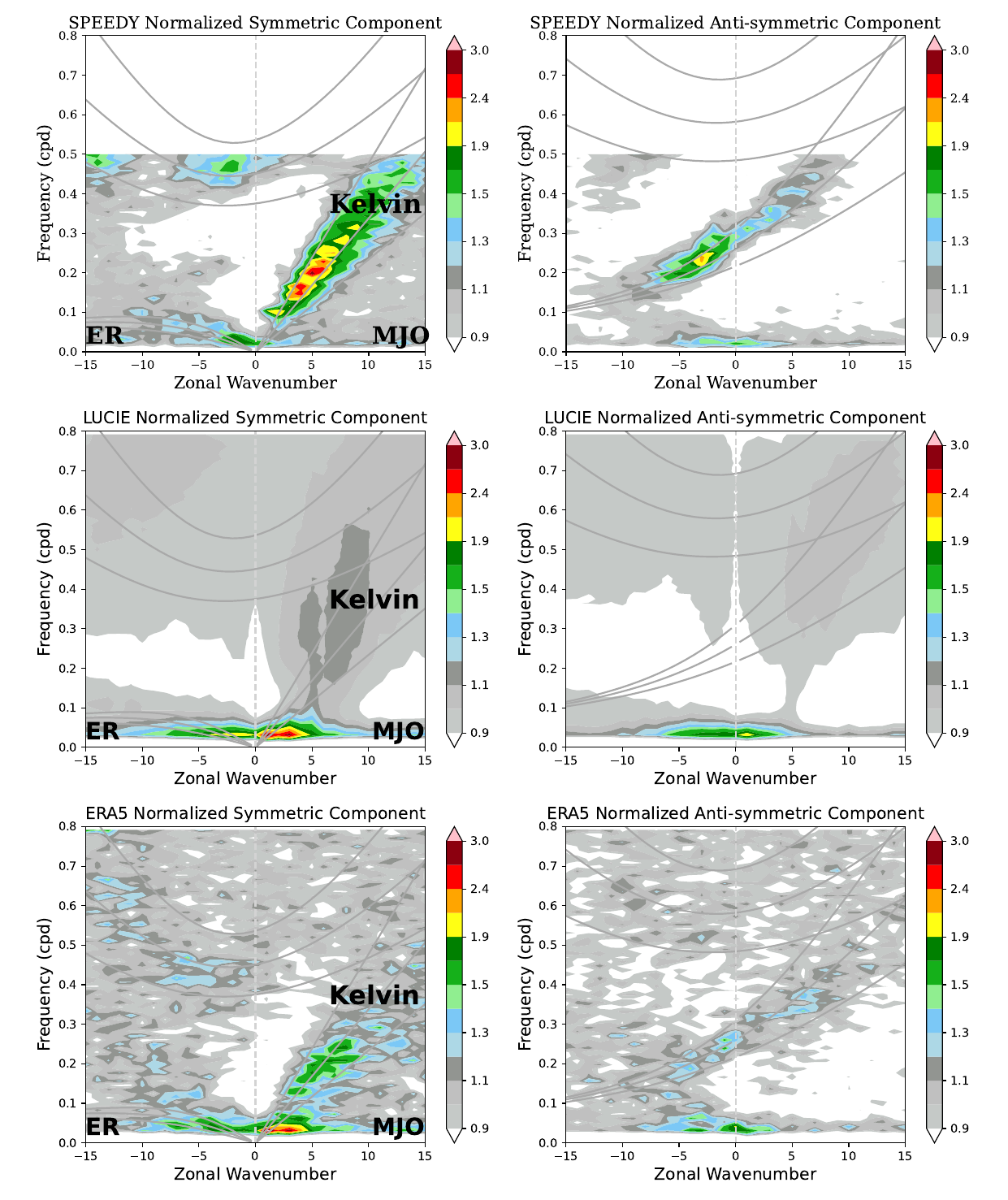}}
\caption{Wheeler-Kiladis diagram of $100$ years of LUCIE inference averaged over 100 ensembles (bottom row) with respect to $10$-year averaged ERA5 and SPEEDY. The diagram of SPEEDY has a maximum frequency of 0.5 because the data from SPEEDY is sampled daily.}
\label{fig:wk_diagram}
\end{center}
\end{figure}

One of the major metrics for physical consistency of a long-term emulator of the climate system is how well the time-space spectra of precipitation is captured in a Wheeler-Kiladis diagram. This is especially important in the context of identifying the Madden-Julian Oscillation (MJO), one of the main modes of variability in the tropics and is particularly important for subseasonal forecasts and a driver of extreme precipitation.  Figure~\ref{fig:wk_diagram} shows that while the spectra of Equatorial Rossby waves (ER) are approximately captured, we do not capture the Kelvin waves. LUCIE is in good agreement with ERA5 for the spectral power in the MJO region (in the lower positive zonal wave numbers and at lower frequencies). SPEEDY is the opposite, where it has very little power where the MJO should be, while overestimating Kelvin waves compared to ERA5. Accommodating the $3$D structure of the humidity field as well as coupling an ocean emulator with sea-surface temperature would likely improve the representation of the spectrum of the Kelvin waves and is attractive for future extensions of the current formulation. It should be noted that in Fig.~\ref{fig:wk_diagram}, the ERA5 data is averaged over $10$ years as compared to LUCIE which is averaged over $100$ years of simulation with $100$ ensembles. As a result, one obtains a smoother spectra for LUCIE. It is also noted that SPEEDY outputs precipitation with a frequency of once per day leading to the top half of Fig.~\ref{fig:wk_diagram} being blank for the panels corresponding to SPEEDY.

\begin{figure}[!h]
%\vskip 0.2in
\begin{center}
\centerline{\includegraphics[width=\textwidth]{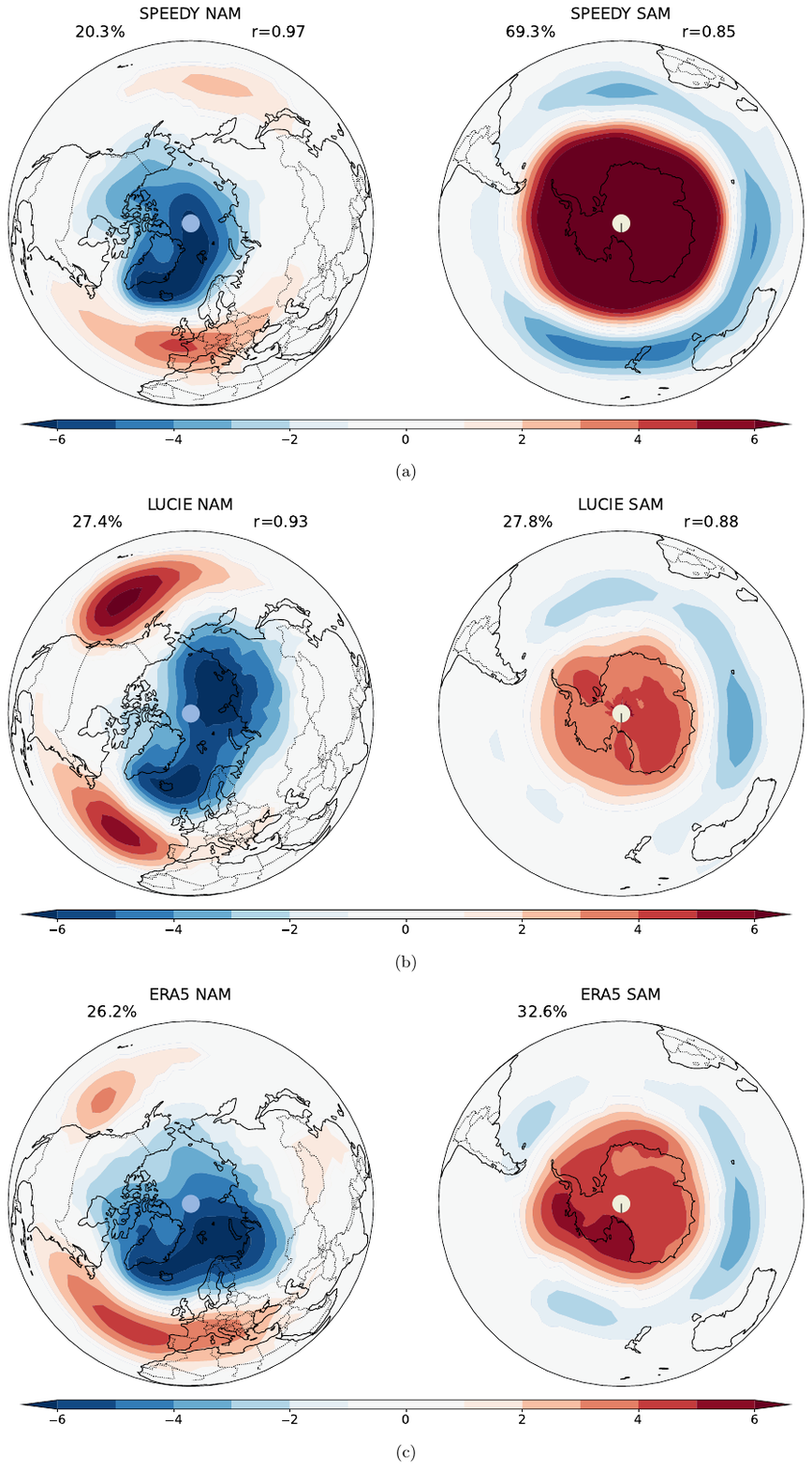}}
\caption{ Northern Hemisphere Annular Mode (NAM) and Southern Hemisphere Annular Mode (SAM) of (a) SEEDY (b) LUCIE and (c) ERA5, with variance fraction and correlation between LUCIE EOF1 and ERA5 EOF1. For LUCIE, $100$ years of simulation have been used to compute the covariance matrices in the EOF analysis while $10$ years of data have been used for both the ERA5 and SPEEDY EOFs. The EOF is calculated with full hemisphere weighted with $\sqrt{\cos(\text{latitude})}$.}
\label{fig:eof}
\end{center}
\end{figure}

Finally, we investigate the first empirical orthogonal function (EOF) of surface pressure over the Northern Hemisphere, i.e.,  Northern Annular Mode (NAM) and Southern Hemisphere, i.e., Southern Annular Mode (SAM) as emulated by LUCIE over $100$ years with the first ensemble member. The EOFs are calculated with surface pressure anomalies for December, January and February. 10 years of data is used for the EOFs of ERA5 and a single ensemble of SPEEDY. Figure~\ref{fig:eof} clearly shows that LUCIE approximately captures the structure of the NAM (correlation coefficient, $r=0.93$), while SAM is represented with slightly lower $r=0.88$. LUCIE correctly captures the percentage of explained variance for both NAM and SAM when compared to ERA5. SPEEDY shows worse performance on both capturing the percentage of variance and the correlation. EOFs are difficult to capture even for simple canonical geophysical systems~\cite{chattopadhyay2023long}. This results is therefore promising for future developments of data-driven emulators of the climate system.

\subsection{LUCIE spectrum}
\begin{figure}[!h]
%\vskip 0.2in
\begin{center}
\centerline{\includegraphics[width=1.1\textwidth]{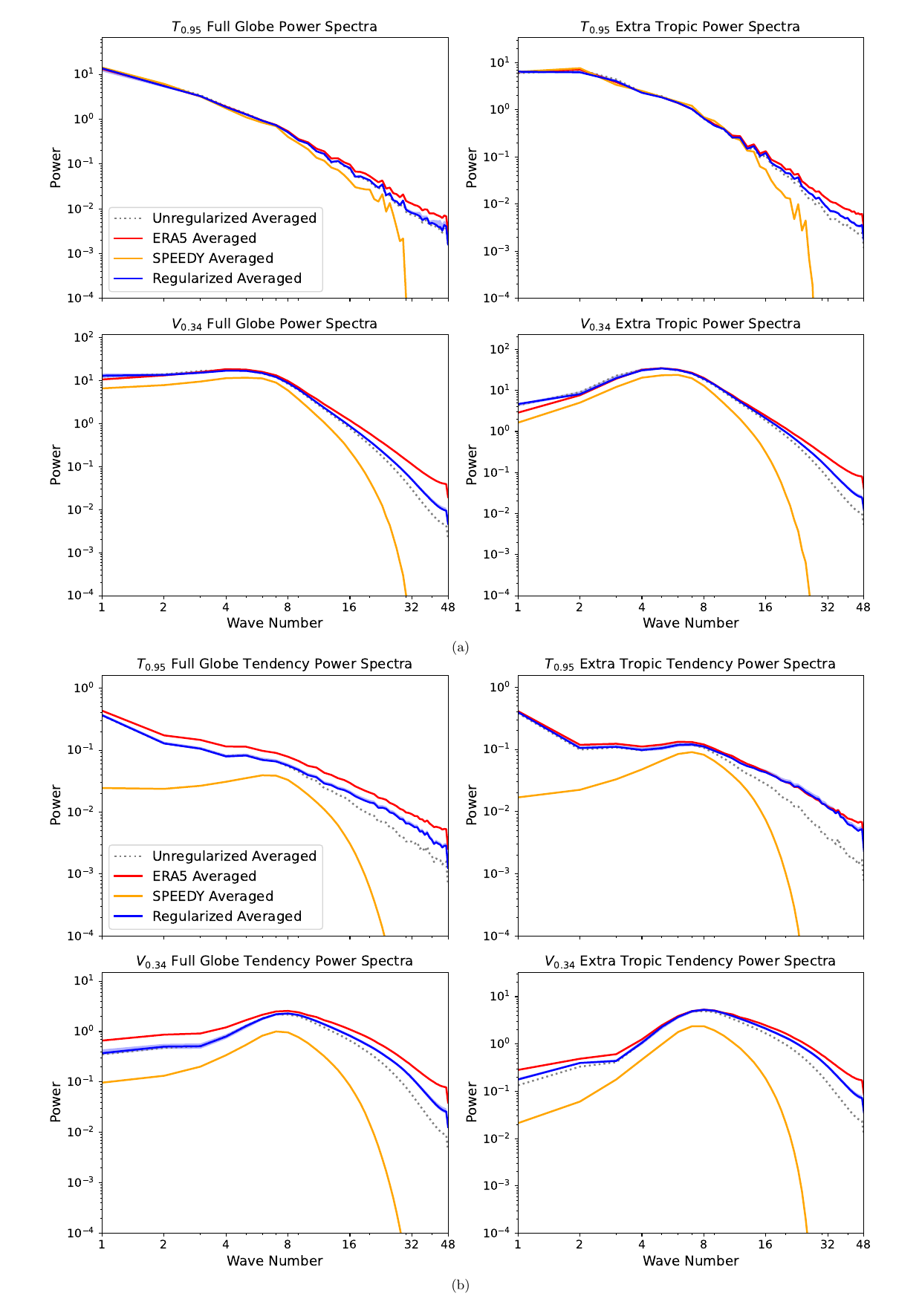}}
\caption{Power spectra of the $T_{0.95}$ and $V_{0.34}$ and their tendencies averaged over a $100$ years of LUCIE simulation over a $100$ ensemble members as compared to ERA5 averaged and SPEEDY over $10$ years. a) Power spectrum of full states. b) Power spectrum of tendency.}
\label{fig:fft_regularizer}
\end{center}
\end{figure}

The major causal mechanism for state-of-the-art AI weather models to go unstable or unphysical is their incorrect representation of the Fourier spectra of key prognostic variables. This has been elucidated theoretically in \citeA{chattopadhyay2023long} and a mitigation strategy had been proposed in the form of a spectral regularizer. A similar strategy has been adopted in this paper, but instead of matching the Fourier spectra of the prognostic states, the spectra of the tendency of the prognostic states near the extra-tropics have been penalized in the regularization term, while the spectrum of the full state is matched for the diagnostic variable. In Fig.~\ref{fig:fft_regularizer} (b), we show that the regularization improves the spectra of the tendency of the variables, $T_{0.95}$, $V_{0.34}$, not just in the extra-tropics (panels of the right) but also over the globe (panels on the left). Fig.~\ref{fig:fft_regularizer} (a) shows the spectra of the full states also improve due to the regularization term imposed on the tendency. Both versions of LUCIE follow closer to ERA5 for all wavenumbers than SPEEDY especially at the highest wavenumbers. Because SPEEDY is a spectral transform model with a spectral resolution of T30, there is no power above wavenumber 30 with the dynamics being heavily filtered starting around wave number 16. SPEEDY also tends to under estimate the power for both components of wind for all wavemumbers Finally, the inclusion of the regularization term improves the overall climatological bias averaged over $100$ years and $100$ ensemble members as shown in Table~\ref{tab:rmse}.

\subsection{Extremes}
\begin{figure}[!h]
%\vskip 0.2in
\begin{center}
\centerline{\includegraphics[width=0.95\textwidth]{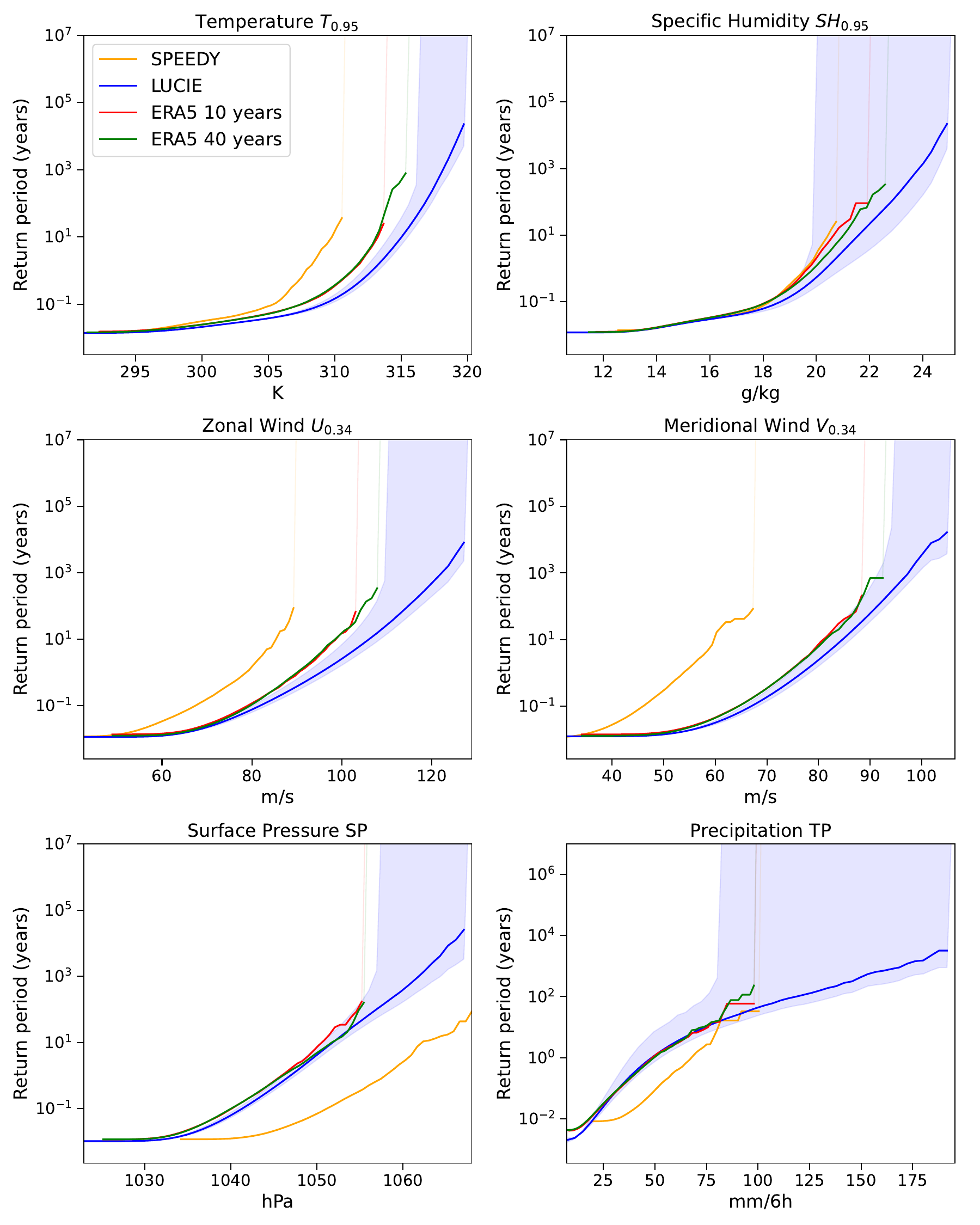}}
\caption{Return period of extreme events computed from $100$ years of LUCIE simulation using $100$ ensemble members and compared to the $10$ year ERA5 training data, $10$ year SPEEDY simulation, and $35$ years of ERA5 validation data.}
\label{fig:return_period}
\end{center}
% \vskip -0.4in
\end{figure}

One of the reasons why data-driven climate emulators are attractive is their ability to use the generated large ensemble of long simulations to reduce sampling error when estimating extreme weather events and their return periods. Moreover, since these models can be run inexpensively for a long time, one may expect to capture extreme events that have return periods longer than the extremes in the training dataset. This is especially useful since our observational records only span $40$ years. However getting the extremes correctly, i.e., accurately estimating the tails of the probability density function (PDF) of the prognostic variables requires that the long-term simulations accurately captures higher-order moments of the PDF, beyond mean and variance (variability) of the climate system. In order to assess how well extremes are captured and how accurate the estimated return periods for such extremes are, we use the generalized extreme value theory to estimate the return period of extreme weather events in the $100$ ensembles of $100$ years simulation and SPEEDY following ~\citeA{emanuel2017assessing}. It should be noted that LUCIE is trained on $10$ years of ERA5 data but integrate for $100$ years. Figure~\ref{fig:return_period} shows that the return period with their uncertainty (estimated as the standard deviation across the ensembles) are overestimated in almost all prognostic variables except for precipitation and surface pressure. In contrast, SPEEDY tends to severely under-predict extreme values for temperature, wind and precipitation. For all variables LUCIE is closer to ERA5 with the exception of specific humidity. We hypothesis this is due to the composite biases in SPEEDY where SPEEDY tends to under-predict extremes, but is too moist for much of the globe. In Fig.~\ref{fig:return_period}, the return period curve for $40$ years of ERA5 is a validation dataset to investigate whether the LUCIE simulation beyond $10$ years realistically captures the return period curve from the validation dataset. In temperature, humidity, and wind, we can see that while the uncertainty bound from LUCIE's simulation does encapsulate the return periods estimated from ERA5, the mean value does not, for the $40$ year period. Hence, the return periods of extremes emulated by LUCIE over $100$ years that are larger than what is found in ERA5 may not be realistic and would require further investigation. For precipitation and surface pressure, we find that the return periods are much better captured within the $40$ year period over which we have ERA5 data. Hence, the extremes estimated by the model beyond $40$ years may be more realistic than the other variables. However for precipitation, we see that the uncertainty grows very quickly for beyond $100$mm/6h values and should not be considered as realistic or physically-consistent events. Generally, LUCIE does produce extremes that are unseen in the training data ($10$ years of ERA5) which other data-driven AI weather models simply cannot~\cite{sun2024can}. In essence, we find that capturing the mean climatology, variability, or EOFs are not enough for a stable climate emulator to simulate realistic grey swan events with the correct return period (i.e., extreme events that are never seen by the model in the training dataset). This remains an open challenge for data-driven climate models that would be developed in the future. Despite this, LUCIE does perform better than SPEEDY for all variables expect for specific humidity. 

\subsection {Performance of LUCIE at low-data regime}
\begin{figure}[!h]
%\vskip 0.2in
\begin{center}
\centerline{\includegraphics[width=0.95\textwidth]{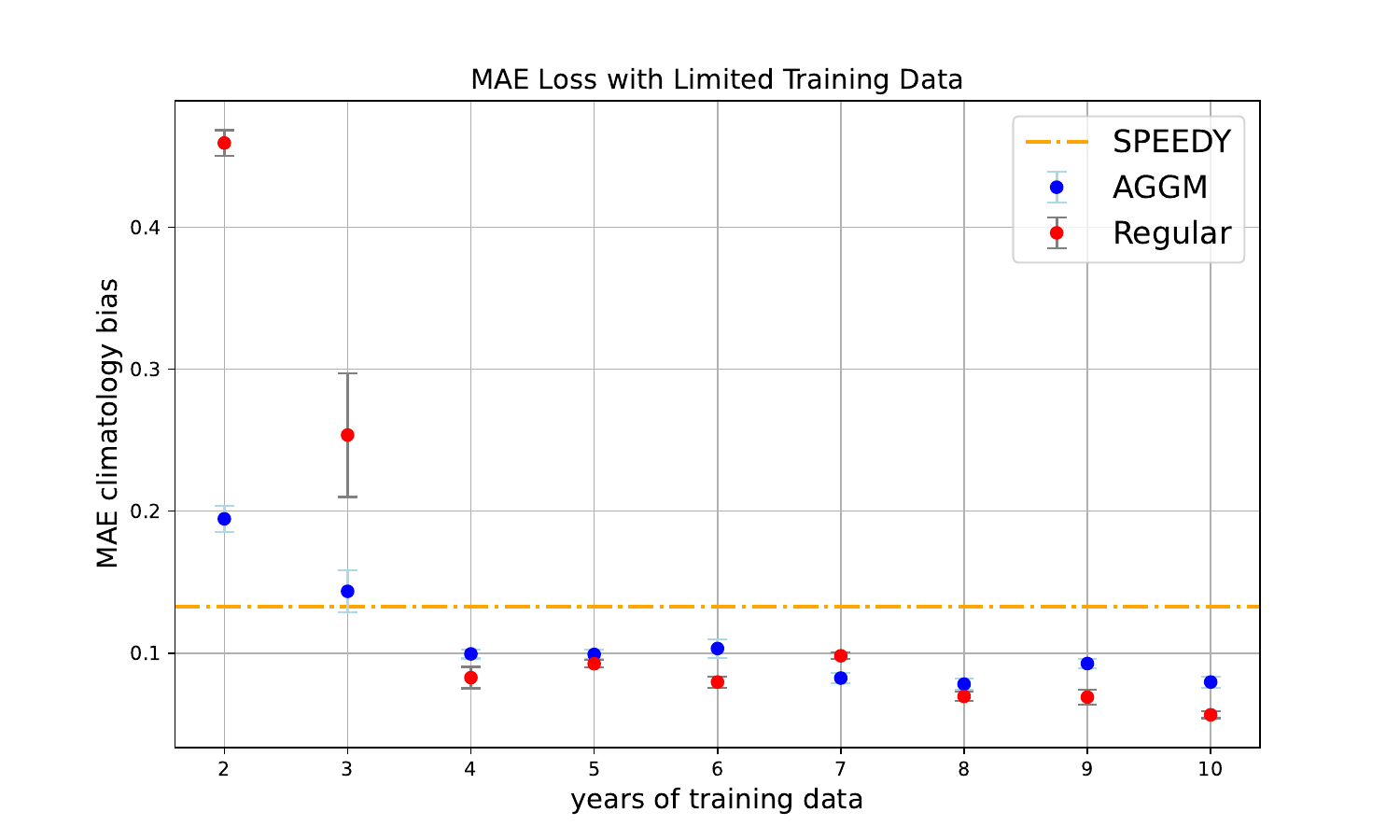}}
\caption{Scaling of LUCIE’s $100$-year climatology (averaged over $100$ ensemble members) with respect to training sample size (in years). The yellow dashed line indicates the MAE of SPEEDY climatology relative to ERA5. Errors are computed using data normalized by the ERA5 mean and standard deviation for readability.}
\label{fig:scale}
\end{center}
\vskip -0.4in
\end{figure}

A question of interest when developing data-driven forecasting models is \textit{how much training data do we minimally require to produce a stable and realistic climate?} We have tried to answer that in this study by training LUCIE on as little as two years of ERA5 data. Previous works in both data-driven weather or in climate models, e.g., ACE, train their models on a large amount of data, e.g., all the state-of-the-art data-driven weather models are trained on roughly $37$ years of data, while ACE has been trained on a $100$ years of climate simulations. In this paper, we train LUCIE on a max of $10$ years of ERA5 data and as little as $2$ years of ERA5 data to investigate how their long-term stability and physical consistency changes. In order to facilitate training with such small sample sizes, we introduced the AGGM method of training at low-data regime that utilizes the memory of computed gradients when updating the parameters of the model during stochastic gradient descent. Figure~\ref{fig:scale} shows that for the low-data regime the AGGM method is stable and shows low climatology bias as compared to regular stochastic gradient descent. Beyond $5$ years of training data both the methods shows roughly similar climatology bias, remains stable, and physically consistent. Both methods of training show that with as little as 3 years of training, LUCIE has lower biases than SPEEDY. The fact that LUCIE can be trained with such low data requirements without drastically affecting its stability and physical consistency makes it a useful model to be used in resource-constrained environments or for rapid prototyping and sandbox experiments. 

\section{Discussion}
\label{sec:diss}

In this paper, we present an efficient lightweight climate emulator, LUCIE, that can autoregressively emulate the atmosphere for $100$ years with $100$ ensemble members. LUCIE is a low resolution model and can be trained on a single GPU for $2.4$h with as less as $2$ years of ERA5 data. LUCIE remains stable and physically consistent with an accurate long-term mean climatology and variability. Overall, LUCIE provides a cheap alternatively to state-of-the-art AI weather models (which are either unstable or physically inconsistent) to conduct in-depth scientific investigations into model performance, scaling, the effect of adding and removing different variables, forced response experiments, as well as novel loss functions and architectural innovation when building climate emulators. 

LUCIE incorporates a first-order integration scheme inside the architecture to suppress autoregressive error growth and also imposes a regularizer on the Fourier spectrum of the tendency of the prognostic variables over the extra-tropics. The inclusion of this regularizer improves the representation of the Fourier spectrum, over the extra-tropics and the globe in both the states and the tendency when integrated for a $100$ years. The spectral regularizer demonstrably shows an improvement in the mean climatology bias. 

While the mean climatology is well represented by LUCIE, it does show deviations from the observed variability in the climate system. While SAM is very accurately captured by the $100$-year LUCIE simulation, the emulated NAM structure deviates from the observed structure, especially when we investigate the pattern correlation of the observed and emulated EOF. Variability in precipitation is accurate in structure but the width of the inter-tropical convergence zone is slightly larger in the simulation than that of the observed during the winter seasons. The time-space spectra also fails to capture the Kelvin waves, although a weak MJO signal is captured. 

We have also investigated how well LUCIE captures the extremes in the prognostic variables. The return periods emulated by LUCIE shows that it significantly overestimates the frequency of extremes in most variables except surface pressure and precipitation. One of the applications of data-driven climate emulators is that they can be run for significantly longer periods of time with large ensembles at a much lower computational cost than climate models. Hence, they may be used to estimate extreme events with tighter uncertainty that are not captured in observational records. LUCIE also predicts extreme events that are higher in intensity with longer return periods than what is observed in the $10$ year ERA5 data on which it has been trained. Whether or not these extreme events are physical remains to be investigated by further analyzing the dynamics of the states that led up to the onset of such extremes. While this paper does not discuss such analysis, an inexpensive model like LUCIE allows us to conduct these studies in future work. 

LUCIE is an uncoupled emulator and does not have the 3D structure of the atmosphere. Inclusion of the humidity structure in more vertical levels can improve the structure of the precipitation time-space spectra. Moreover, inclusion of a slab ocean emulator to couple sea-surface temperature dynamics is an immediate next step for LUCIE as we continue to improve the model without significant addition to computational cost. 

Lastly, we note that while the training data used for LUCIE contains a nonstationary climate, during long free runs LUCIE produces a stable and stationary climate. This is an undesirable feature for an emulator if the purpose is to study climate change. Further research and development is needed before climate emulators such as the one presented in this study can be used for nonstationary climates. Future research directions to improve the generalization of these models to different climates may include the addition of atmospheric chemistry (e.g. greenhouse gases), longer training datasets to better capture the climate change signal, and the utilization of both observation-based analyses and model simulation of future climate scenarios.

\clearpage
\newpage

\appendix

\section{Hyperparameters}

Table ~\ref{tab:hyperparameters} lists the SFNO hyperparameters used in LUCIE. 
\begin{table}[h]
\centering
\begin{tabular}{ll}
\toprule
\textbf{Hyperparameter} & \textbf{Value} \\
\midrule
SFNO blocks & \num{8} \\
Encoder \& decoder layers & \num{1} \\
Latent dimension & \num{72} \\
Maximum Learning Rate & \num{1e-4}\\
Minimum Learning Rate & \num{1e-6}\\
Batch Size & \num{16} \\
Number of Epochs & \num{250} \\
Optimizer & Adam \\
Activation Function & SiLU \\
Regularizer weight & 5e-2 \\
Weight perturbation STD & 5e-4 \\
Precipitation episilon & 1e-2\\
\bottomrule
\end{tabular}
\vskip 0.1in
\caption{SFNO model hyperparameters.}
\label{tab:hyperparameters}
\end{table}

\section{AGGM training with 2 years training data}
\label{sec:aggm_appendix}

\begin{figure}[htb]
    \centering
    \begin{minipage}{0.48\textwidth} % Allocate half the text width to each minipage
        \begin{algorithm}[H]
        \caption{Batch gradient descent}
        \begin{algorithmic}[1] % The number tells where the line numbering should start
        \State Initialize model parameters $\theta$
        \State Define loss function $\mathcal{L}$
        \For{each epoch}
            \For{each batch $b$ in training data}
                \State Zero gradients: $\nabla \theta \gets 0$
                \State Forward pass: Compute predicted outputs $\hat{y}_b$
                \State Compute loss: $\ell_b \gets \mathcal{L}(\hat{y}_b, y_b)$
                \State Backward pass: Compute gradient $\nabla \ell_b$
                \State Update parameters: $\theta \gets \theta - \eta \nabla \ell_b$
            \EndFor
        \EndFor
        \end{algorithmic}
        \end{algorithm}
    \end{minipage}\hfill
    \begin{minipage}{0.48\textwidth}
        \begin{algorithm}[H]
        \caption{AGGM}
        \begin{algorithmic}[1] % The number tells where the line numbering should start
        \State Initialize model parameters $\theta$
        \State Define loss function $\mathcal{L}$
        \For{each epoch}
            \State Zero gradients: $\nabla \theta \gets 0$
            \For{each batch $b$ in training data}
                \State Forward pass: Compute predicted outputs $\hat{y}_b$
                \State Compute loss: $\ell_b \gets \mathcal{L}(\hat{y}_b, y_b)$
                \State Backward pass: Compute gradient $\nabla \ell_b$
                \State Update parameters: $\theta \gets \theta - \eta \nabla \ell_b$
            \EndFor
        \EndFor
        \end{algorithmic}
        \end{algorithm}
    \end{minipage}
\end{figure}

\begin{figure}[!h]
%\vskip 0.2in
\begin{center}
\centerline{\includegraphics[width=0.9\textwidth]{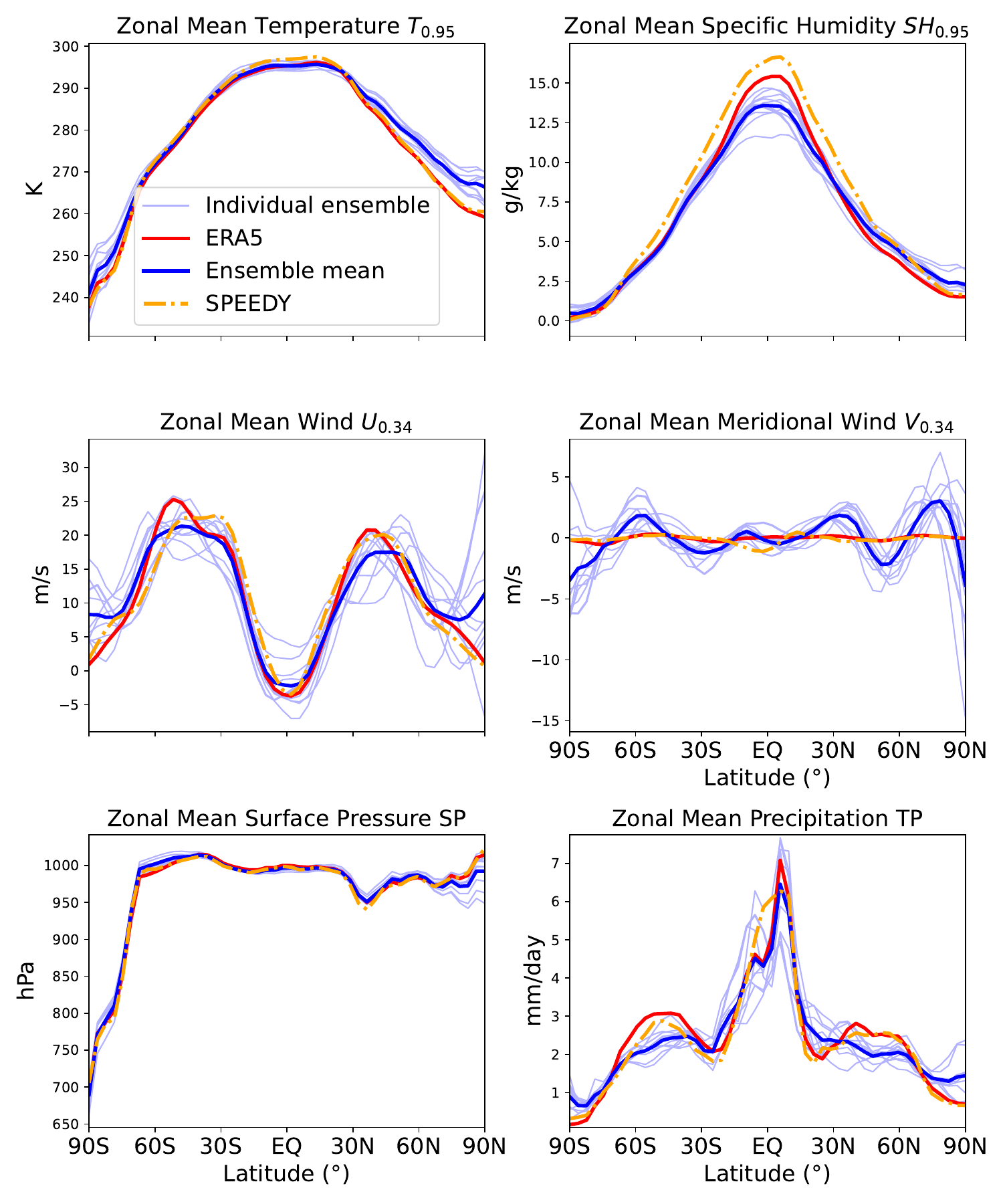}}
\caption{Zonal mean climatology of LUCIE with AGGM trained with 2 years of ERA5. The climatology is averaged over 10 years of inference with 10 ensmebles.}
\label{fig:aggm_clim}
\end{center}
% \vskip -0.4in
\end{figure}

To demonstrate the capabilities of AGGM, we present the 100-year zonal mean climatology produced by LUCIE using AGGM, trained with two years of ERA5 data (see ~\ref{fig:aggm_clim}). Although the climatology bias is larger compared to training with ten years of data, LUCIE maintains long-term stability while accurately capturing long-term statistics. This ability enables rapid experimentation with various techniques, such as adjustments to loss functions and hyperparameter tuning. While further investigation is needed to evaluate the generalizability of this method to other datasets, it shows promising potential for reducing data requirements in data-driven climate model.

\clearpage
\newpage

\section{Power Spectrum}

\begin{figure}[!h]
%\vskip 0.2in
\begin{center}
\centerline{\includegraphics[width=0.9\textwidth]{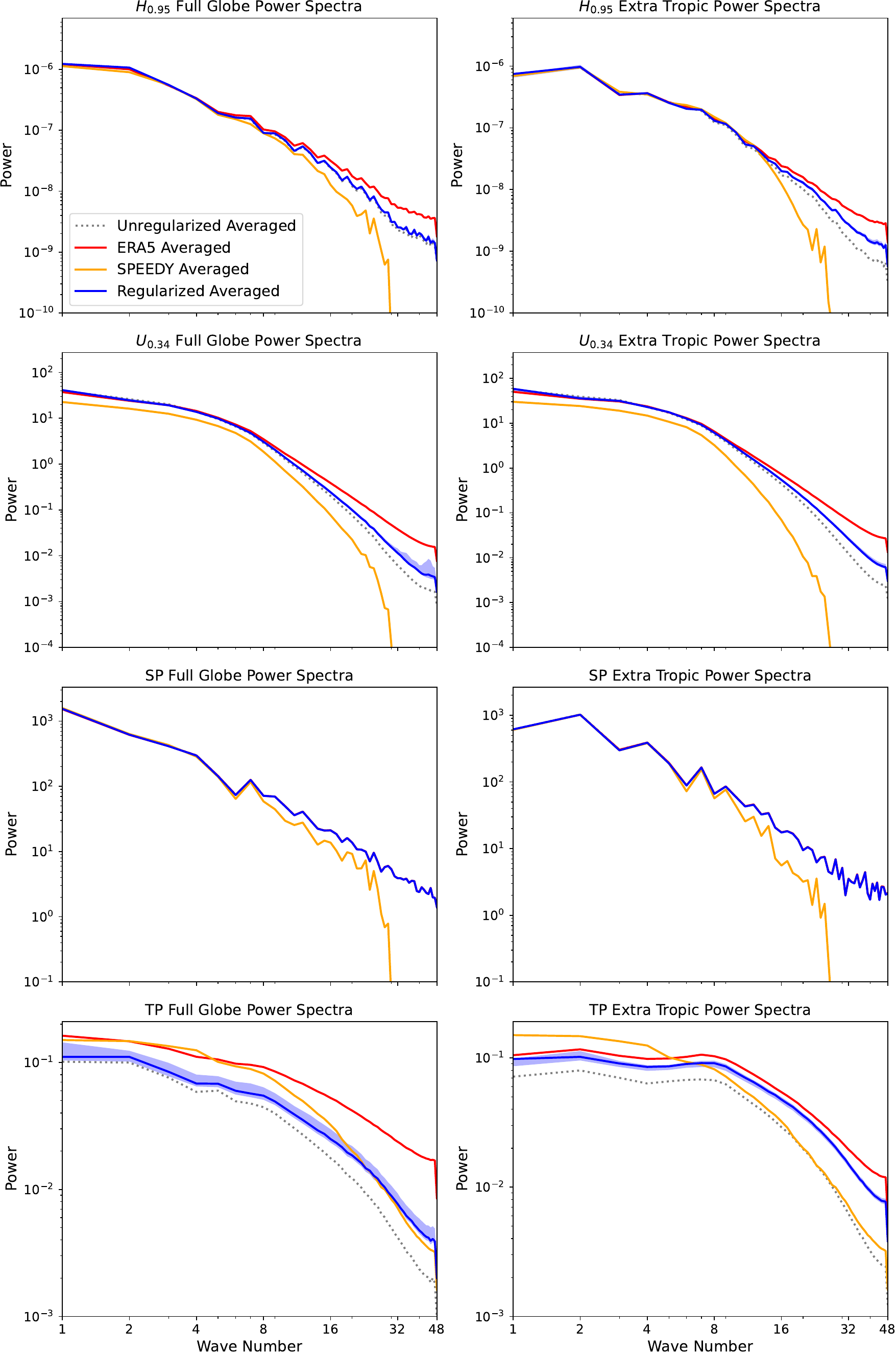}}
\caption{Full state power spectrum of humidity, zonal wind, surface pressure, precipitation.}
\label{fig:other_field_spectrum}
\end{center}
% \vskip -0.4in
\end{figure}

\begin{figure}[!h]
%\vskip 0.2in
\begin{center}
\centerline{\includegraphics[width=0.9\textwidth]{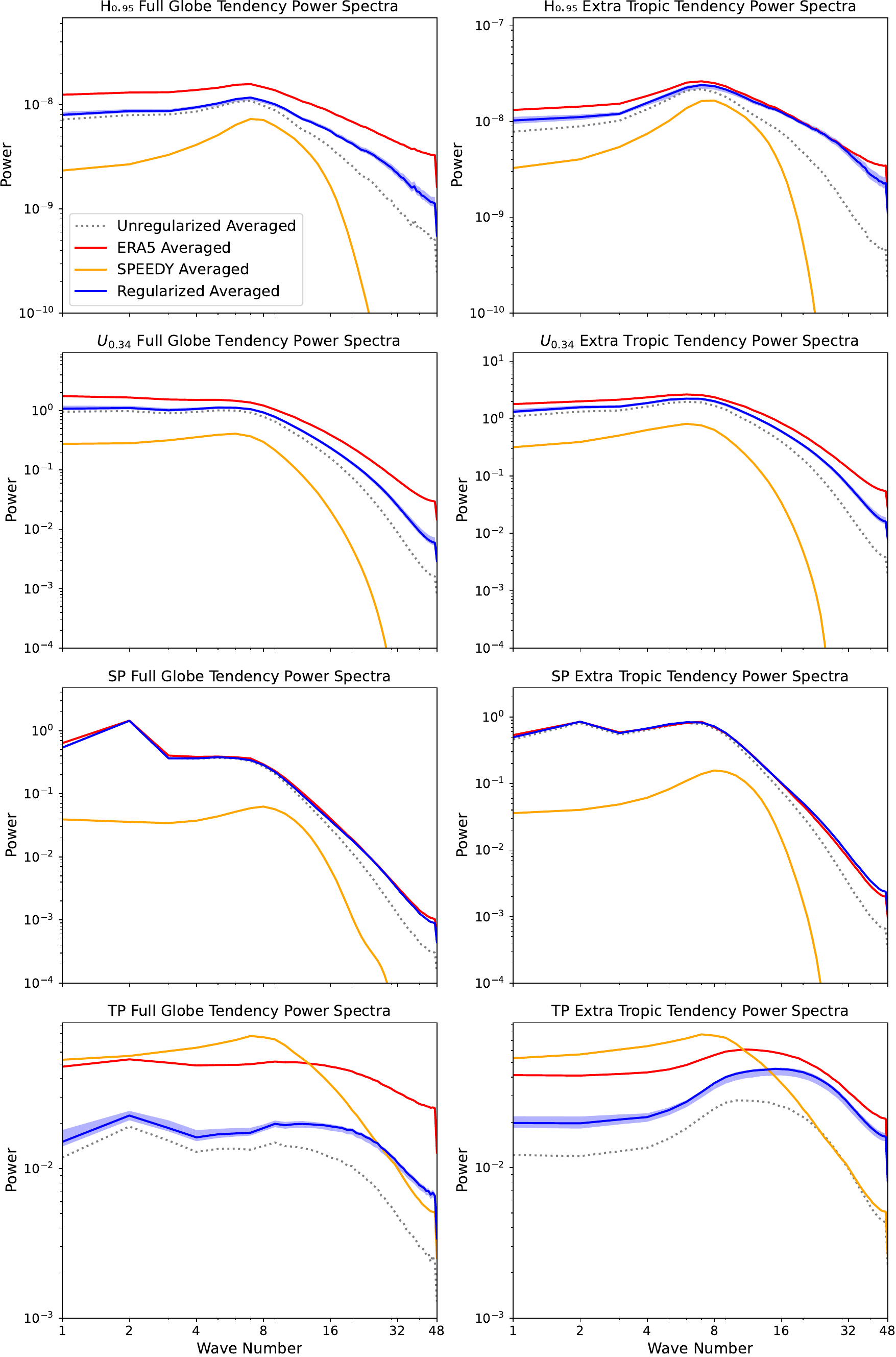}}
\caption{Tendency power spectrum of humidity, zonal wind, surface pressure, precipitation.}
\label{fig:other_tendency_spectrum}
\end{center}
% \vskip -0.4in
\end{figure}

This section provides the FFT power spectrum plots for variables not included in Fig. ~\ref{fig:fft_regularizer}. The full-field spectrum is shown in Table~\ref{fig:other_field_spectrum}, while the spectrum of the variable tendencies is shown in Table~\ref{fig:other_tendency_spectrum}.

\newpage

\section*{Open Research Section}
The codes used for training and inference are open hosted on Github \\
(https://github.com/ISCLPennState/LUCIE.git) and also permanently archived on Zenodo (https://zenodo.org/records/15164648). The regridded ERA5 data for $10$ years needed for training and the trained weights can be found in the same Zenodo link. 
% The trained weights for LUCIE will be made available online upon publication. 

\acknowledgments

This material is based upon work supported by the U.S. Department of Energy (DOE), Office of Science, Office of Advanced Scientific Computing Research, under Contract DE-AC02-06CH11357. This research was funded in part and used resources of the Argonne Leadership Computing Facility, which is a DOE Office of Science User Facility supported under Contract DE-AC02-06CH11357. RM acknowledges support from U.S. Department of Energy, Office of Science, Office of Advanced Scientific Computing Research grant DOE-FOA-2493: ``Data-intensive scientific machine learning'', and computational support from Penn-State Institute for Computational and Data Sciences. TA was supported
by the Global Change Fellowship in the Environmental Science Division at Argonne National Laboratory (grant no.
LDRD 2023-0236). HG was partially supported by Argonne National Laboratory (grant no.
LDRD 2023-0236). AC was supported by the National Science Foundation (grant no. 2425667) and computational support from NSF ACCESS MTH240019 and NCAR CISL UCSC0008. 
%%%%%%%%%%%%%%%%%%%%%%%%%%%%%%%%%%%%%%%%%%%%%%%
% REFERENCES and BIBLIOGRAPHY
%
% \bibliography{<name of your .bib file>} don't specify the file extension
% don't specify bibliographystyle
%
%%%%%%%%%%%%%%%%%%%%%%%%%%%%%%%%%%%%%%%%%%%%%%%

\bibliography{agusample}

\begin{thebibliography}{}

\bibitem [\protect \citeauthoryear {%
Adamov%
, Schemm%
, Fuhrer%
\BCBL {}\ \BBA {} Knutti%
}{%
Adamov%
\ \protect \BOthers {.}}{%
{\protect \APACyear {2024}}%
}]{%
pasc2024pos122}
\APACinsertmetastar {%
pasc2024pos122}%
\begin{APACrefauthors}%
Adamov, S.%
, Schemm, S.%
, Fuhrer, O.%
\BCBL {}\ \BBA {} Knutti, R.%
\end{APACrefauthors}%
\unskip\
\newblock
\APACrefYearMonthDay{2024}{}{}.
\newblock
\APACrefbtitle {Probabilistic Weather Forecasting through Latent Space Perturbations of Machine Learning Emulators.} {Probabilistic weather forecasting through latent space perturbations of machine learning emulators.}
\newblock
\APAChowpublished {Poster presented at PASC24 Conference, Session 158}.
\newblock
\APACrefnote{Available at: \url{https://pasc24.pasc-conference.org/presentation/?id=pos122&sess=sess158}}
\PrintBackRefs{\CurrentBib}

\bibitem [\protect \citeauthoryear {%
Arcomano%
\ \protect \BOthers {.}}{%
Arcomano%
\ \protect \BOthers {.}}{%
{\protect \APACyear {2022}}%
}]{%
Arcomano_2022}
\APACinsertmetastar {%
Arcomano_2022}%
\begin{APACrefauthors}%
Arcomano, T.%
, Szunyogh, I.%
, Wikner, A.%
, Pathak, J.%
, Hunt, B\BPBI R.%
\BCBL {}\ \BBA {} Ott, E.%
\end{APACrefauthors}%
\unskip\
\newblock
\APACrefYearMonthDay{2022}{}{}.
\newblock
{\BBOQ}\APACrefatitle {A Hybrid Approach to Atmospheric Modeling That Combines Machine Learning With a Physics-Based Numerical Model} {A hybrid approach to atmospheric modeling that combines machine learning with a physics-based numerical model}.{\BBCQ}
\newblock
\APACjournalVolNumPages{Journal of Advances in Modeling Earth Systems}{14}{3}{e2021MS002712}.
\newblock
\begin{APACrefDOI} \doi{https://doi.org/10.1029/2021MS002712} \end{APACrefDOI}
\PrintBackRefs{\CurrentBib}

\bibitem [\protect \citeauthoryear {%
Bi%
\ \protect \BOthers {.}}{%
Bi%
\ \protect \BOthers {.}}{%
{\protect \APACyear {2023}}%
}]{%
bi2023accurate}
\APACinsertmetastar {%
bi2023accurate}%
\begin{APACrefauthors}%
Bi, K.%
, Xie, L.%
, Zhang, H.%
, Chen, X.%
, Gu, X.%
\BCBL {}\ \BBA {} Tian, Q.%
\end{APACrefauthors}%
\unskip\
\newblock
\APACrefYearMonthDay{2023}{}{}.
\newblock
{\BBOQ}\APACrefatitle {Accurate medium-range global weather forecasting with {3D} neural networks} {Accurate medium-range global weather forecasting with {3D} neural networks}.{\BBCQ}
\newblock
\APACjournalVolNumPages{Nature}{619}{7970}{533--538}.
\PrintBackRefs{\CurrentBib}

\bibitem [\protect \citeauthoryear {%
Bloch-Johnson%
\ \protect \BOthers {.}}{%
Bloch-Johnson%
\ \protect \BOthers {.}}{%
{\protect \APACyear {2024}}%
}]{%
bloch2024green}
\APACinsertmetastar {%
bloch2024green}%
\begin{APACrefauthors}%
Bloch-Johnson, J.%
, Rugenstein, M\BPBI A.%
, Alessi, M\BPBI J.%
, Proistosescu, C.%
, Zhao, M.%
, Zhang, B.%
\BDBL {}others%
\end{APACrefauthors}%
\unskip\
\newblock
\APACrefYearMonthDay{2024}{}{}.
\newblock
{\BBOQ}\APACrefatitle {The green's function model intercomparison project (GFMIP) protocol} {The green's function model intercomparison project (gfmip) protocol}.{\BBCQ}
\newblock
\APACjournalVolNumPages{Journal of Advances in Modeling Earth Systems}{16}{2}{e2023MS003700}.
\PrintBackRefs{\CurrentBib}

\bibitem [\protect \citeauthoryear {%
Bonavita%
}{%
Bonavita%
}{%
{\protect \APACyear {2023}}%
}]{%
bonavita2023limitations}
\APACinsertmetastar {%
bonavita2023limitations}%
\begin{APACrefauthors}%
Bonavita, M.%
\end{APACrefauthors}%
\unskip\
\newblock
\APACrefYearMonthDay{2023}{}{}.
\newblock
{\BBOQ}\APACrefatitle {On the limitations of data-driven weather forecasting models} {On the limitations of data-driven weather forecasting models}.{\BBCQ}
\newblock
\APACjournalVolNumPages{arXiv preprint arXiv:2309.08473}{}{}{}.
\PrintBackRefs{\CurrentBib}

\bibitem [\protect \citeauthoryear {%
Bonavita%
}{%
Bonavita%
}{%
{\protect \APACyear {2024}}%
}]{%
Massimo2024}
\APACinsertmetastar {%
Massimo2024}%
\begin{APACrefauthors}%
Bonavita, M.%
\end{APACrefauthors}%
\unskip\
\newblock
\APACrefYearMonthDay{2024}{2024/09/22}{}.
\newblock
{\BBOQ}\APACrefatitle {On Some Limitations of Current Machine Learning Weather Prediction Models} {On some limitations of current machine learning weather prediction models}.{\BBCQ}
\newblock
\APACjournalVolNumPages{Geophysical Research Letters}{51}{12}{e2023GL107377}.
\newblock
\begin{APACrefURL} \url{https://doi.org/10.1029/2023GL107377} \end{APACrefURL}
\newblock
\begin{APACrefDOI} \doi{https://doi.org/10.1029/2023GL107377} \end{APACrefDOI}
\PrintBackRefs{\CurrentBib}

\bibitem [\protect \citeauthoryear {%
Bonev%
\ \protect \BOthers {.}}{%
Bonev%
\ \protect \BOthers {.}}{%
{\protect \APACyear {2023}}%
}]{%
bonev2023spherical}
\APACinsertmetastar {%
bonev2023spherical}%
\begin{APACrefauthors}%
Bonev, B.%
, Kurth, T.%
, Hundt, C.%
, Pathak, J.%
, Baust, M.%
, Kashinath, K.%
\BCBL {}\ \BBA {} Anandkumar, A.%
\end{APACrefauthors}%
\unskip\
\newblock
\APACrefYearMonthDay{2023}{}{}.
\newblock
{\BBOQ}\APACrefatitle {Spherical fourier neural operators: Learning stable dynamics on the sphere} {Spherical fourier neural operators: Learning stable dynamics on the sphere}.{\BBCQ}
\newblock
\BIn{} \APACrefbtitle {International conference on machine learning} {International conference on machine learning}\ (\BPGS\ 2806--2823).
\PrintBackRefs{\CurrentBib}

\bibitem [\protect \citeauthoryear {%
Chattopadhyay%
, Gray%
, Wu%
, Lowe%
\BCBL {}\ \BBA {} He%
}{%
Chattopadhyay%
\ \protect \BOthers {.}}{%
{\protect \APACyear {2023}}%
}]{%
chattopadhyay2023oceannet}
\APACinsertmetastar {%
chattopadhyay2023oceannet}%
\begin{APACrefauthors}%
Chattopadhyay, A.%
, Gray, M.%
, Wu, T.%
, Lowe, A\BPBI B.%
\BCBL {}\ \BBA {} He, R.%
\end{APACrefauthors}%
\unskip\
\newblock
\APACrefYearMonthDay{2023}{}{}.
\newblock
{\BBOQ}\APACrefatitle {OceanNet: A principled neural operator-based digital twin for regional oceans} {Oceannet: A principled neural operator-based digital twin for regional oceans}.{\BBCQ}
\newblock
\APACjournalVolNumPages{arXiv preprint arXiv:2310.00813}{}{}{}.
\PrintBackRefs{\CurrentBib}

\bibitem [\protect \citeauthoryear {%
Chattopadhyay%
\ \BBA {} Hassanzadeh%
}{%
Chattopadhyay%
\ \BBA {} Hassanzadeh%
}{%
{\protect \APACyear {2023}}%
}]{%
chattopadhyay2023long}
\APACinsertmetastar {%
chattopadhyay2023long}%
\begin{APACrefauthors}%
Chattopadhyay, A.%
\BCBT {}\ \BBA {} Hassanzadeh, P.%
\end{APACrefauthors}%
\unskip\
\newblock
\APACrefYearMonthDay{2023}{}{}.
\newblock
{\BBOQ}\APACrefatitle {Long-term instabilities of deep learning-based digital twins of the climate system: The cause and a solution} {Long-term instabilities of deep learning-based digital twins of the climate system: The cause and a solution}.{\BBCQ}
\newblock
\APACjournalVolNumPages{arXiv preprint arXiv:2304.07029}{}{}{}.
\PrintBackRefs{\CurrentBib}

\bibitem [\protect \citeauthoryear {%
Driscoll%
\ \BBA {} Healy%
}{%
Driscoll%
\ \BBA {} Healy%
}{%
{\protect \APACyear {1994}}%
}]{%
driscoll1994}
\APACinsertmetastar {%
driscoll1994}%
\begin{APACrefauthors}%
Driscoll, J\BPBI R.%
\BCBT {}\ \BBA {} Healy, D\BPBI M.%
\end{APACrefauthors}%
\unskip\
\newblock
\APACrefYearMonthDay{1994}{}{}.
\newblock
{\BBOQ}\APACrefatitle {Computing Fourier Transforms and Convolutions on the 2-Sphere} {Computing fourier transforms and convolutions on the 2-sphere}.{\BBCQ}
\newblock
\APACjournalVolNumPages{Advances in Applied Mathematics}{15}{2}{202--250}.
\newblock
\begin{APACrefURL} \url{https://www.sciencedirect.com/science/article/pii/S0196885884710086} \end{APACrefURL}
\newblock
\begin{APACrefDOI} \doi{https://doi.org/10.1006/aama.1994.1008} \end{APACrefDOI}
\PrintBackRefs{\CurrentBib}

\bibitem [\protect \citeauthoryear {%
Duncan%
\ \protect \BOthers {.}}{%
Duncan%
\ \protect \BOthers {.}}{%
{\protect \APACyear {2024}}%
}]{%
Duncan2024}
\APACinsertmetastar {%
Duncan2024}%
\begin{APACrefauthors}%
Duncan, J\BPBI P\BPBI C.%
, Wu, E.%
, Golaz, J\BHBI C.%
, Caldwell, P\BPBI M.%
, Watt-Meyer, O.%
, Clark, S\BPBI K.%
\BDBL {}Bretherton, C\BPBI S.%
\end{APACrefauthors}%
\unskip\
\newblock
\APACrefYearMonthDay{2024}{2024/09/15}{}.
\newblock
{\BBOQ}\APACrefatitle {Application of the AI2 Climate Emulator to E3SMv2's Global Atmosphere Model, With a Focus on Precipitation Fidelity} {Application of the ai2 climate emulator to e3smv2's global atmosphere model, with a focus on precipitation fidelity}.{\BBCQ}
\newblock
\APACjournalVolNumPages{Journal of Geophysical Research: Machine Learning and Computation}{1}{3}{e2024JH000136}.
\newblock
\begin{APACrefURL} \url{https://doi.org/10.1029/2024JH000136} \end{APACrefURL}
\newblock
\begin{APACrefDOI} \doi{https://doi.org/10.1029/2024JH000136} \end{APACrefDOI}
\PrintBackRefs{\CurrentBib}

\bibitem [\protect \citeauthoryear {%
Emanuel%
}{%
Emanuel%
}{%
{\protect \APACyear {2017}}%
}]{%
emanuel2017assessing}
\APACinsertmetastar {%
emanuel2017assessing}%
\begin{APACrefauthors}%
Emanuel, K.%
\end{APACrefauthors}%
\unskip\
\newblock
\APACrefYearMonthDay{2017}{}{}.
\newblock
{\BBOQ}\APACrefatitle {Assessing the present and future probability of Hurricane Harvey’s rainfall} {Assessing the present and future probability of hurricane harvey’s rainfall}.{\BBCQ}
\newblock
\APACjournalVolNumPages{Proceedings of the National Academy of Sciences}{114}{48}{12681--12684}.
\PrintBackRefs{\CurrentBib}

\bibitem [\protect \citeauthoryear {%
Hakim%
\ \BBA {} Masanam%
}{%
Hakim%
\ \BBA {} Masanam%
}{%
{\protect \APACyear {2023}}%
}]{%
hakim2023dynamical}
\APACinsertmetastar {%
hakim2023dynamical}%
\begin{APACrefauthors}%
Hakim, G\BPBI J.%
\BCBT {}\ \BBA {} Masanam, S.%
\end{APACrefauthors}%
\unskip\
\newblock
\APACrefYearMonthDay{2023}{}{}.
\newblock
{\BBOQ}\APACrefatitle {Dynamical tests of a deep-learning weather prediction model} {Dynamical tests of a deep-learning weather prediction model}.{\BBCQ}
\newblock
\APACjournalVolNumPages{arXiv preprint arXiv:2309.10867}{}{}{}.
\PrintBackRefs{\CurrentBib}

\bibitem [\protect \citeauthoryear {%
Hersbach%
\ \protect \BOthers {.}}{%
Hersbach%
\ \protect \BOthers {.}}{%
{\protect \APACyear {2020}}%
}]{%
Hersbach2020}
\APACinsertmetastar {%
Hersbach2020}%
\begin{APACrefauthors}%
Hersbach, H.%
, Bell, B.%
, Berrisford, P.%
, Hirahara, S.%
, Horányi, A.%
, Muñoz-Sabater, J.%
\BDBL {}Thépaut, J\BHBI N.%
\end{APACrefauthors}%
\unskip\
\newblock
\APACrefYearMonthDay{2020}{}{}.
\newblock
{\BBOQ}\APACrefatitle {The {ERA5} global reanalysis} {The {ERA5} global reanalysis}.{\BBCQ}
\newblock
\APACjournalVolNumPages{Quarterly Journal of the Royal Meteorological Society}{146}{730}{1999-2049}.
\newblock
\begin{APACrefDOI} \doi{https://doi.org/10.1002/qj.3803} \end{APACrefDOI}
\PrintBackRefs{\CurrentBib}

\bibitem [\protect \citeauthoryear {%
Karlbauer%
\ \protect \BOthers {.}}{%
Karlbauer%
\ \protect \BOthers {.}}{%
{\protect \APACyear {2024}}%
}]{%
karlbauer2024advancing}
\APACinsertmetastar {%
karlbauer2024advancing}%
\begin{APACrefauthors}%
Karlbauer, M.%
, Cresswell-Clay, N.%
, Durran, D\BPBI R.%
, Moreno, R\BPBI A.%
, Kurth, T.%
, Bonev, B.%
\BDBL {}Butz, M\BPBI V.%
\end{APACrefauthors}%
\unskip\
\newblock
\APACrefYearMonthDay{2024}{}{}.
\newblock
{\BBOQ}\APACrefatitle {Advancing parsimonious deep learning weather prediction using the HEALPix mesh} {Advancing parsimonious deep learning weather prediction using the healpix mesh}.{\BBCQ}
\newblock
\APACjournalVolNumPages{Journal of Advances in Modeling Earth Systems}{16}{8}{e2023MS004021}.
\PrintBackRefs{\CurrentBib}

\bibitem [\protect \citeauthoryear {%
Keisler%
}{%
Keisler%
}{%
{\protect \APACyear {2022}}%
}]{%
keisler2022forecasting}
\APACinsertmetastar {%
keisler2022forecasting}%
\begin{APACrefauthors}%
Keisler, R.%
\end{APACrefauthors}%
\unskip\
\newblock
\APACrefYearMonthDay{2022}{}{}.
\newblock
{\BBOQ}\APACrefatitle {Forecasting global weather with graph neural networks} {Forecasting global weather with graph neural networks}.{\BBCQ}
\newblock
\APACjournalVolNumPages{arXiv preprint arXiv:2202.07575}{}{}{}.
\PrintBackRefs{\CurrentBib}

\bibitem [\protect \citeauthoryear {%
Kochkov%
\ \protect \BOthers {.}}{%
Kochkov%
\ \protect \BOthers {.}}{%
{\protect \APACyear {2024}}%
}]{%
kochkov2024neural}
\APACinsertmetastar {%
kochkov2024neural}%
\begin{APACrefauthors}%
Kochkov, D.%
, Yuval, J.%
, Langmore, I.%
, Norgaard, P.%
, Smith, J.%
, Mooers, G.%
\BDBL {}others%
\end{APACrefauthors}%
\unskip\
\newblock
\APACrefYearMonthDay{2024}{}{}.
\newblock
{\BBOQ}\APACrefatitle {Neural general circulation models for weather and climate} {Neural general circulation models for weather and climate}.{\BBCQ}
\newblock
\APACjournalVolNumPages{Nature}{}{}{1--7}.
\PrintBackRefs{\CurrentBib}

\bibitem [\protect \citeauthoryear {%
Kucharski%
, Molteni%
\BCBL {}\ \BBA {} Bracco%
}{%
Kucharski%
\ \protect \BOthers {.}}{%
{\protect \APACyear {2006}}%
}]{%
Kucharski2006}
\APACinsertmetastar {%
Kucharski2006}%
\begin{APACrefauthors}%
Kucharski, F.%
, Molteni, F.%
\BCBL {}\ \BBA {} Bracco, A.%
\end{APACrefauthors}%
\unskip\
\newblock
\APACrefYearMonthDay{2006}{}{}.
\newblock
{\BBOQ}\APACrefatitle {Decadal interactions between the western tropical Pacific and the North Atlantic Oscillation} {Decadal interactions between the western tropical pacific and the north atlantic oscillation}.{\BBCQ}
\newblock
\APACjournalVolNumPages{Climate Dynamics}{26}{1}{79-91}.
\PrintBackRefs{\CurrentBib}

\bibitem [\protect \citeauthoryear {%
Lam%
\ \protect \BOthers {.}}{%
Lam%
\ \protect \BOthers {.}}{%
{\protect \APACyear {2023}}%
}]{%
lam2022graphcast}
\APACinsertmetastar {%
lam2022graphcast}%
\begin{APACrefauthors}%
Lam, R.%
, Sanchez-Gonzalez, A.%
, Willson, M.%
, Wirnsberger, P.%
, Fortunato, M.%
, Alet, F.%
\BDBL {}Battaglia, P.%
\end{APACrefauthors}%
\unskip\
\newblock
\APACrefYearMonthDay{2023}{}{}.
\newblock
{\BBOQ}\APACrefatitle {Learning skillful medium-range global weather forecasting} {Learning skillful medium-range global weather forecasting}.{\BBCQ}
\newblock
\APACjournalVolNumPages{Science}{0}{0}{eadi2336}.
\newblock
\begin{APACrefURL} \url{https://www.science.org/doi/abs/10.1126/science.adi2336} \end{APACrefURL}
\newblock
\begin{APACrefDOI} \doi{10.1126/science.adi2336} \end{APACrefDOI}
\PrintBackRefs{\CurrentBib}

\bibitem [\protect \citeauthoryear {%
Molteni%
}{%
Molteni%
}{%
{\protect \APACyear {2003}}%
}]{%
molteni2003}
\APACinsertmetastar {%
molteni2003}%
\begin{APACrefauthors}%
Molteni, F.%
\end{APACrefauthors}%
\unskip\
\newblock
\APACrefYearMonthDay{2003}{}{}.
\newblock
{\BBOQ}\APACrefatitle {Atmospheric simulations using a {GCM} with simplified physical parametrizations. {I}: model climatology and variability in multi-decadal experiments} {Atmospheric simulations using a {GCM} with simplified physical parametrizations. {I}: model climatology and variability in multi-decadal experiments}.{\BBCQ}
\newblock
\APACjournalVolNumPages{Climate Dynamics}{20}{2}{175-191}.
\PrintBackRefs{\CurrentBib}

\bibitem [\protect \citeauthoryear {%
Nguyen%
\ \protect \BOthers {.}}{%
Nguyen%
\ \protect \BOthers {.}}{%
{\protect \APACyear {2024}}%
}]{%
nguyen2024scaling}
\APACinsertmetastar {%
nguyen2024scaling}%
\begin{APACrefauthors}%
Nguyen, T.%
, Shah, R.%
, Bansal, H.%
, Arcomano, T.%
, Madireddy, S.%
, Maulik, R.%
\BDBL {}Grover, A.%
\end{APACrefauthors}%
\unskip\
\newblock
\APACrefYearMonthDay{2024}{}{}.
\newblock
{\BBOQ}\APACrefatitle {Scaling Transformers for Skillful and Reliable Medium-range Weather Forecasting} {Scaling transformers for skillful and reliable medium-range weather forecasting}.{\BBCQ}
\newblock
\BIn{} \APACrefbtitle {ICLR 2024 Workshop on Tackling Climate Change with Machine Learning (Best Paper).} {Iclr 2024 workshop on tackling climate change with machine learning (best paper).}
\PrintBackRefs{\CurrentBib}

\bibitem [\protect \citeauthoryear {%
Patel%
, Arcomano%
, Hunt%
, Szunyogh%
\BCBL {}\ \BBA {} Ott%
}{%
Patel%
\ \protect \BOthers {.}}{%
{\protect \APACyear {2024}}%
}]{%
patel2024exploring}
\APACinsertmetastar {%
patel2024exploring}%
\begin{APACrefauthors}%
Patel, D.%
, Arcomano, T.%
, Hunt, B.%
, Szunyogh, I.%
\BCBL {}\ \BBA {} Ott, E.%
\end{APACrefauthors}%
\unskip\
\newblock
\APACrefYearMonthDay{2024}{}{}.
\newblock
{\BBOQ}\APACrefatitle {Exploring the Potential of Hybrid Machine-Learning/Physics-Based Modeling for Atmospheric/Oceanic Prediction Beyond the Medium Range} {Exploring the potential of hybrid machine-learning/physics-based modeling for atmospheric/oceanic prediction beyond the medium range}.{\BBCQ}
\newblock
\APACjournalVolNumPages{arXiv preprint arXiv:2405.19518}{}{}{}.
\PrintBackRefs{\CurrentBib}

\bibitem [\protect \citeauthoryear {%
Pathak%
\ \protect \BOthers {.}}{%
Pathak%
\ \protect \BOthers {.}}{%
{\protect \APACyear {2022}}%
}]{%
pathak2022fourcastnet}
\APACinsertmetastar {%
pathak2022fourcastnet}%
\begin{APACrefauthors}%
Pathak, J.%
, Subramanian, S.%
, Harrington, P.%
, Raja, S.%
, Chattopadhyay, A.%
, Mardani, M.%
\BDBL {}Anandkumar, A.%
\end{APACrefauthors}%
\unskip\
\newblock
\APACrefYearMonthDay{2022}{}{}.
\newblock
{\BBOQ}\APACrefatitle {Four{CastNet}: A global data-driven high-resolution weather model using adaptive {F}ourier neural operators} {Four{CastNet}: A global data-driven high-resolution weather model using adaptive {F}ourier neural operators}.{\BBCQ}
\newblock
\APACjournalVolNumPages{arXiv preprint arXiv:2202.11214}{}{}{}.
\PrintBackRefs{\CurrentBib}

\bibitem [\protect \citeauthoryear {%
Price%
\ \protect \BOthers {.}}{%
Price%
\ \protect \BOthers {.}}{%
{\protect \APACyear {2023}}%
}]{%
price2023gencast}
\APACinsertmetastar {%
price2023gencast}%
\begin{APACrefauthors}%
Price, I.%
, Sanchez-Gonzalez, A.%
, Alet, F.%
, Ewalds, T.%
, El-Kadi, A.%
, Stott, J.%
\BDBL {}Willson, M.%
\end{APACrefauthors}%
\unskip\
\newblock
\APACrefYearMonthDay{2023}{}{}.
\newblock
{\BBOQ}\APACrefatitle {Gencast: Diffusion-based ensemble forecasting for medium-range weather} {Gencast: Diffusion-based ensemble forecasting for medium-range weather}.{\BBCQ}
\newblock
\APACjournalVolNumPages{arXiv preprint arXiv:2312.15796}{}{}{}.
\PrintBackRefs{\CurrentBib}

\bibitem [\protect \citeauthoryear {%
Subich%
}{%
Subich%
}{%
{\protect \APACyear {2024}}%
}]{%
subich2024efficient}
\APACinsertmetastar {%
subich2024efficient}%
\begin{APACrefauthors}%
Subich, C.%
\end{APACrefauthors}%
\unskip\
\newblock
\APACrefYearMonthDay{2024}{}{}.
\newblock
{\BBOQ}\APACrefatitle {Efficient fine-tuning of 37-level GraphCast with the Canadian global deterministic analysis} {Efficient fine-tuning of 37-level graphcast with the canadian global deterministic analysis}.{\BBCQ}
\newblock
\APACjournalVolNumPages{arXiv preprint arXiv:2408.14587}{}{}{}.
\PrintBackRefs{\CurrentBib}

\bibitem [\protect \citeauthoryear {%
Sun%
\ \protect \BOthers {.}}{%
Sun%
\ \protect \BOthers {.}}{%
{\protect \APACyear {2024}}%
}]{%
sun2024can}
\APACinsertmetastar {%
sun2024can}%
\begin{APACrefauthors}%
Sun, Y\BPBI Q.%
, Hassanzadeh, P.%
, Zand, M.%
, Chattopadhyay, A.%
, Weare, J.%
\BCBL {}\ \BBA {} Abbot, D\BPBI S.%
\end{APACrefauthors}%
\unskip\
\newblock
\APACrefYearMonthDay{2024}{}{}.
\newblock
{\BBOQ}\APACrefatitle {Can AI weather models predict out-of-distribution gray swan tropical cyclones?} {Can ai weather models predict out-of-distribution gray swan tropical cyclones?}{\BBCQ}
\newblock
\APACjournalVolNumPages{arXiv preprint arXiv:2410.14932}{}{}{}.
\PrintBackRefs{\CurrentBib}

\bibitem [\protect \citeauthoryear {%
Watt-Meyer%
\ \protect \BOthers {.}}{%
Watt-Meyer%
\ \protect \BOthers {.}}{%
{\protect \APACyear {2021}}%
}]{%
watt2021correcting}
\APACinsertmetastar {%
watt2021correcting}%
\begin{APACrefauthors}%
Watt-Meyer, O.%
, Brenowitz, N\BPBI D.%
, Clark, S\BPBI K.%
, Henn, B.%
, Kwa, A.%
, McGibbon, J.%
\BDBL {}Bretherton, C\BPBI S.%
\end{APACrefauthors}%
\unskip\
\newblock
\APACrefYearMonthDay{2021}{}{}.
\newblock
{\BBOQ}\APACrefatitle {Correcting weather and climate models by machine learning nudged historical simulations} {Correcting weather and climate models by machine learning nudged historical simulations}.{\BBCQ}
\newblock
\APACjournalVolNumPages{Geophysical Research Letters}{48}{15}{e2021GL092555}.
\PrintBackRefs{\CurrentBib}

\bibitem [\protect \citeauthoryear {%
Watt-Meyer%
\ \protect \BOthers {.}}{%
Watt-Meyer%
\ \protect \BOthers {.}}{%
{\protect \APACyear {2023}}%
}]{%
watt2023ace}
\APACinsertmetastar {%
watt2023ace}%
\begin{APACrefauthors}%
Watt-Meyer, O.%
, Dresdner, G.%
, McGibbon, J.%
, Clark, S\BPBI K.%
, Henn, B.%
, Duncan, J.%
\BDBL {}others%
\end{APACrefauthors}%
\unskip\
\newblock
\APACrefYearMonthDay{2023}{}{}.
\newblock
{\BBOQ}\APACrefatitle {ACE: A fast, skillful learned global atmospheric model for climate prediction} {Ace: A fast, skillful learned global atmospheric model for climate prediction}.{\BBCQ}
\newblock
\APACjournalVolNumPages{arXiv preprint arXiv:2310.02074}{}{}{}.
\PrintBackRefs{\CurrentBib}

\end{thebibliography}

%Reference citation instructions and examples:
%
% Please use ONLY \cite and \citeA for reference citations.
% \cite for parenthetical references
% ...as shown in recent studies (Simpson et al., 2019)
% \citeA for in-text citations
% ...Simpson et al. (2019) have shown...
%
%
%...as shown by \citeA{jskilby}.
%...as shown by \citeA{lewin76}, \citeA{carson86}, \citeA{bartoldy02}, and \citeA{rinaldi03}.
%...has been shown \cite{jskilbye}.
%...has been shown \cite{lewin76,carson86,bartoldy02,rinaldi03}.
%... \cite <i.e.>[]{lewin76,carson86,bartoldy02,rinaldi03}.
%...has been shown by \cite <e.g.,>[and others]{lewin76}.
%
% apacite uses < > for prenotes and [ ] for postnotes
% DO NOT use other cite commands (e.g., \citet, \citep, \citeyear, \nocite, \citealp, etc.).
%

\end{document}